\documentclass[runningheads]{llncs}
\usepackage{graphicx}
\usepackage[table]{xcolor}
\usepackage{tikz}
\usepackage{comment}
\usepackage{amsmath,amssymb} % define this before the line numbering.
\usepackage{color}
\usepackage{multirow}
\usepackage{gensymb}
\usepackage{pifont}
\usepackage{enumerate}

\usepackage{graphicx}
\usepackage{booktabs}
\usepackage{microtype}      % microtypography
\usepackage{textcomp}
\usepackage{pifont}
\usepackage{soul}
\usepackage{mathtools}
\usepackage{bm}
\usepackage{paralist, tabularx}

\usepackage[accsupp]{axessibility}  % Improves PDF readability for those with disabilities.

\definecolor{citecolor}{HTML}{0071bc}
\usepackage[pagebackref=true,breaklinks=true,letterpaper=true,colorlinks,citecolor=citecolor,bookmarks=false]{hyperref}

\newcommand\red[1]{\textcolor{red}{\textsuperscript{#1}}}

\newcommand{\cmark}{\ding{51}}%
\newcommand{\xmark}{\ding{55}}%
\newcommand{\greencmark}{\textcolor{green}{\cmark}}
\newcommand{\redxmark}{\textcolor{red}{\xmark}}
\definecolor{hlrowcolor}{rgb}{1.0,0.9,0.8}

\newcommand{\ie}{\textit{i}.\textit{e}., }
\newcommand{\eg}{\textit{e}.\textit{g}., }

\usepackage{array}

\DeclareMathOperator*{\argmax}{arg\,max}
\DeclareMathOperator*{\argmin}{arg\,min}

\def\*#1{{\bm{#1}}}

\usepackage[export]{adjustbox}

\usepackage[capitalize]{cleveref}
\crefname{section}{Sec.}{Secs.}
\Crefname{section}{Section}{Sections}
\Crefname{table}{Table}{Tables}
\crefname{table}{Tab.}{Tabs.}

\begin{document}
\pagestyle{headings}
\mainmatter
\def\ECCVSubNumber{4596}  % Insert your submission number here

\title{Cross-Modal 3D Shape Generation and Manipulation} % Replace with your title

% INITIAL SUBMISSION 
\begin{comment}
\titlerunning{ECCV-22 submission ID 4596} 
\authorrunning{ECCV-22 submission ID 4596} 
\author{Anonymous ECCV submission}
\institute{Paper ID 4596}
\end{comment}
%******************

% CAMERA READY SUBMISSION
%\begin{comment}
\titlerunning{Cross-Modal 3D Shape Generation and Manipulation}
% If the paper title is too long for the running head, you can set
% an abbreviated paper title here
%
\author{Zezhou Cheng\inst{1}\thanks{This work was mainly done while the first author was an intern at Snap Inc. Code and data are available at \url{https://people.cs.umass.edu/\~zezhoucheng/edit3d/}} \and
Menglei Chai\inst{2} \and
Jian Ren\inst{2} \and
Hsin-Ying Lee\inst{2} \and
Kyle Olszewski\inst{2} \and
Zeng Huang\inst{2} \and
Subhransu Maji\inst{1} \and
Sergey Tulyakov\inst{2}
}
\institute{University of Massachusetts, Amherst \and Snap Inc.}

\authorrunning{Z. Cheng et al.}
% First names are abbreviated in the running head.
% If there are more than two authors, 'et al.' is used.
%
%\end{comment}
%******************
\maketitle

\begin{abstract}
Creating and editing the shape and color of 3D objects require tremendous human effort and expertise. Compared to direct manipulation in 3D interfaces, 2D interactions such as sketches and scribbles are usually much more natural and intuitive for the users. In this paper, we propose a generic multi-modal generative model that couples the 2D modalities and implicit 3D representations through shared latent spaces. With the proposed model, versatile 3D generation and manipulation are enabled by simply propagating the editing from a specific 2D controlling modality through the latent spaces. For example, editing the 3D shape by drawing a sketch, re-colorizing the 3D surface via painting color scribbles on the 2D rendering, or generating 3D shapes of a certain category given one or a few reference images. Unlike prior works, our model does not require re-training or fine-tuning per editing task and is also conceptually simple, easy to implement, robust to input domain shifts, and flexible to diverse reconstruction on partial 2D inputs. We evaluate our framework on two representative 2D modalities of grayscale line sketches and rendered color images, and demonstrate that our method enables various shape manipulation and generation tasks with these 2D modalities.
\end{abstract}

%%%%%%%%% BODY TEXT
\section{Introduction}
\label{sec:intro}
With the growth in 3D acquisition and visualization technology, there is an increasing need of tools for 3D content creation and editing tasks such as deforming the shape of an object, changing the color of a part, or inserting or removing a component. 
The graphics and vision community has proposed a number of tools for these tasks~\cite{schmidt2016state,delanoy20183d,an2011appwarp,pellacini2007lighting}. Yet, manipulating 3D still requires tremendous human labor and expertise, prohibiting wide-scale adoption by non-professionals.
Compared to the traditional 3D user interfaces, 2D interactions on view-dependent image planes can be a more intuitive way to edit the shape. This has motivated the community to leverage advances in shape representations using deep networks~\cite{park2019deepsdf,mescheder2019occupancy,chen2019learning,sitzmann2019scene} for 3D shape manipulation with 2D controls, such as mesh reconstruction from sketches~\cite{guillard2021sketch2mesh} and color editing with scribbles~\cite{liu2021editing}.
However, most prior works on 2D-to-3D shape manipulation are tailored to a particular editing task and interaction format, which makes generalization to new editing tasks or controls challenging, or even infeasible. 
This is important because there is often no single interaction that fits every use case -- the preferred 2D user control depends on the editing goals, scenarios, devices, or targeted users. 

\begin{figure}[t]
    \centering
    \includegraphics[width=0.94\linewidth]{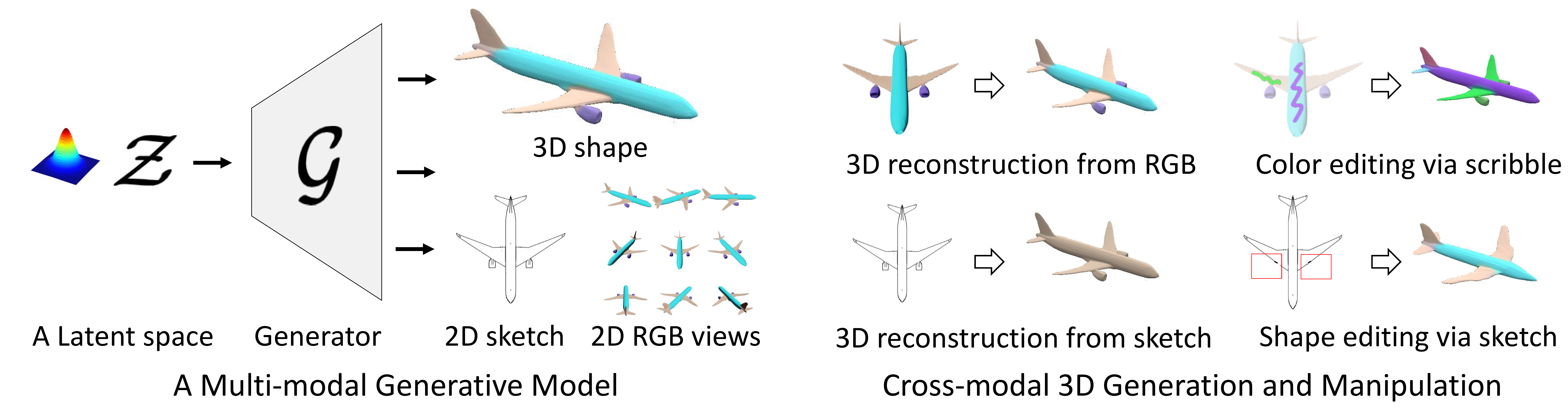}
    \caption{We propose a multi-modal generative model that bridges multiple 2D (\eg sketch, color views) and 3D modalities via shared latent spaces (\emph{left}). Versatile 3D shape generation and manipulation tasks can be tackled via simple latent optimization method (\emph{right}).}
    \label{fig:splash}
\end{figure}

Motivated by this, we propose a 2D-to-3D framework that not only works on a single control modality but also enjoys the flexibility of handling various types of 2D interactions without the need for changing the architecture or even re-training (Fig.~\ref{fig:splash} left).
Our framework bridges various 2D interaction modalities and the target 3D shape through a uniform editing propagation mechanism.
The key is to construct a shared latent representation across generative models of each of the 2D and 3D modalities.
The shared latent representation enforces that an arbitrary latent code corresponds to a 3D model that is consistent with every modality, in terms of both shape and color. With our model, any editing can be achieved by an objective that aims to match the corresponding editing modality and backpropagating the error to estimate the latent code. Moreover, different editing operations and modalities can be combined and interleaved leading to a versatile tool for editing the shape (Fig.~\ref{fig:splash} right).
The approach can be extended to a new user control by simply adding a generator for the corresponding modality in the framework. 

We evaluate our framework on two representative 2D modalities, \ie grayscale line sketches, and rendered color images. We provide extensive quantitative and qualitative results in shape and color editing with sketches and scribbles, as well as single-view, few-shot, or even partial-view cross-modal shape generation. The proposed method is conceptually simple, easy to implement, robust to input domain shifts, and generalizable to new modalities with no special requirement on the network architecture.

\section{Related Work}
\label{sec:related}

\begin{table}[ht!]
   \centering
    \setlength{\tabcolsep}{7pt}
       \caption{\label{tab:difference} \textbf{Comparisons to cross-modal 3D editing and generation works.} }
    \begin{tabular}{cccccc}
    \toprule
    \multirow{2}{*}{Methods} & \multicolumn{2}{c}{\emph{Manipulation}} & \multicolumn{3}{c}{ \emph{Generation}} \\ 
   & Shape & Color & Single view & Partial view & Few shot\\
   \midrule
    Sketch2Mesh~\cite{guillard2021sketch2mesh}  &  \greencmark & \redxmark & \greencmark & \redxmark & \redxmark \\
    DualSDF~\cite{hao2020dualsdf} &  \greencmark & \redxmark &  \redxmark & \redxmark & \redxmark  \\
    EditNeRF~\cite{liu2021editing} &  \greencmark  &  \greencmark  &  \redxmark & \redxmark & \redxmark  \\
    \rowcolor{hlrowcolor}
    Ours &  \greencmark &  \greencmark  &  \greencmark  & \greencmark & \greencmark\\
    \bottomrule
  \end{tabular}
\end{table}

\noindent\textbf{Multi-Modal Generative Models.} 
There has been much work on learning a joint distribution of multiple modalities $p(\*x_0,\dots,\*x_n)$ where each modality $\*x_i$ represents one representation (\eg images, text) of underlying signals. 
Multi-modal VAEs~\cite{suzuki2016joint,wu2018multimodal,wu2019multimodal,shi2019variational,kingma2013auto} learn a joint distribution $p_\theta(\*x_0,\dots,\*x_n\mid\*z)$ conditioned on common latent variables $\*z \in \mathcal{Z}$.
Without the assumption of paired multi-modal data, multi-modal GANs~\cite{liu2016coupled,choi2018stargan,goodfellow2014generative} learn the joint distribution by sharing a latent space and model parameters across modalities. 
These multi-modal generative models have enabled versatile applications such as cross-modal image translation~\cite{choi2018stargan,liu2016coupled} and domain adaptation~\cite{liu2016coupled}. 
Similar to these works, we build a multi-modal generative model that bridges multiple modalities via a shared latent space. 
However, we generate and edit 3D shapes with sparse 2D inputs (\eg scribbles, sketches) and build a 2D-3D generative model based on variational auto-decoders (VADs)~\cite{zadeh2019variational,hao2020dualsdf}. 
Prior work~\cite{zadeh2019variational} has shown that VADs excel at generative modeling from incomplete data. In this work, we demonstrate that the multi-modal VADs (MM-VADs) are ideally suited for the task of 3D generation and manipulation from sparse 2D inputs (\eg color scribble or partial inputs).

\noindent\textbf{Shape and Appearance Reconstruction.} Extensive works have explored the problem of 3D reconstruction from different modalities, such as RGB images~\cite{kanazawa2018learning,choy20163d}, videos~\cite{yang2021lasr}, sketches~\cite{jin2020contour,guillard2021sketch2mesh,zhong2020deep,zhang2021sketch2model}, or even text~\cite{chen2018text2shape}. 
This problem has also been explored under diverse representations~\cite{choy20163d,gkioxari2019mesh,liu2019soft,wang2018pixel2mesh,fan2017point,park2019deepsdf,chen2019learning,mescheder2019occupancy,sitzmann2019scene,xu2019disn} and different levels of supervision~\cite{choy20163d,gkioxari2019mesh,kanazawa2018learning,goel2020shape,yang2021lasr}. 
Despite the diverse settings of this problem, the encoder-decoder network, which maps the source modalities to 3D shape directly in a feed-forward manner, remains the most popular 3D reconstruction model~\cite{choy20163d,wang2018pixel2mesh,kanazawa2018learning,park2019deepsdf}. 
However, such feed-forward networks are not robust to input domain shift (\eg incomplete data). 
In this work, we demonstrate that the proposed MM-VADs perform more robustly and could provide multiple 3D reconstructions that fit the given input (\eg partial 2D views).

\noindent\textbf{Shape and Appearance Manipulation.} 
Numerous interactive tools have been developed for image editing~\cite{levin2004colorization,zhang2017real,rother2004grabcut,li2004lazy,grady2006random,lempitsky2009image} and 3D shape manipulations~\cite{schmidt2016state,delanoy20183d,an2011appwarp,pellacini2007lighting}.
More recently, generative modeling of natural images \cite{goodfellow2014generative,sohl2015deep} has became a ``Swiss knife" for image editing problems~\cite{zhu2020domain,shen2020interfacegan,shen2020interpreting,gu2020image,abdal2019image2stylegan,bau2020rewriting,bau2020semantic,pan2020exploiting,saharia2021palette}.
Similar to these works, we build a multi-modal generative model that is able to tackle versatile 3D shape generation and editing tasks with 2D inputs. 
Novel interactive tools have also been proposed recently to edit implicit 3D representations~\cite{park2019deepsdf,mildenhall2020nerf}. 
For example, DualSDF~\cite{hao2020dualsdf} edits the SDFs~\cite{park2019deepsdf} via shape primitives (\eg spheres). 
Sketch2Mesh~\cite{guillard2021sketch2mesh} reconstructs shapes from sketch with an encoder-decoder network 
and refines 3D shapes via differentiable rendering. 
EditNeRF\cite{liu2021editing} edits the radiance field~\cite{mildenhall2020nerf} by fine-tuning the network weights based on user's scribbles.

Tab.~\ref{tab:difference} summarizes the commons and differences between our work and recent efforts~\cite{hao2020dualsdf,guillard2021sketch2mesh,liu2021editing} on 3D manipulation and generation. 
Similar to Sketch2Mesh~\cite{guillard2021sketch2mesh}, we edit and reconstruct 3D shape from 2D sketch. However, we tackle this problem via a novel multi-modal \emph{generative} model that performs more robust to input domain shift (\eg partial input, sparse color scribble). Furthermore, the shape and color edits can be combined and interleaved with our model; Like EditNeRF, we edit the appearance of 3D shapes via 2D color scribbles. However, we conduct the 3D editing via a simple latent optimization, instead of finetuning the network weights per edit; Akin to DualSDF~\cite{hao2020dualsdf}, we build a generative model for 3D manipulation, yet we generate and edit shapes from 2D modalities which is more intuitive to edit the shape than using 3D primitives. Moreover, our generative model can be adapted to generate 3D shapes of a certain category (\eg armchairs) given a few 2D examples, namely, \emph{few-shot cross-modal shape generation}.  

\section{Method}
\label{sec:method}

We describe the Variational Auto-Decoders (VADs)~\cite{zadeh2019variational} in \S~\ref{subset:VAD}, introduce the proposed VAD-based multi-modal generative model (dubbed MM-VADs) in \S~\ref{subsec:MM-VAD}, and illustrate the application of MM-VADs in cross-modal 3D shape generation and manipulation tasks in \S~\ref{sec:application}.

\subsection{Background: Variational Auto-Decoder}
\label{subset:VAD}

Given observation variables $\*x\sim p(\*x)$ and latent variables $\*z\sim p(\*z)$, a variational auto-decoder (VAD) approximates the data distribution $p(\*x)$ via a parametric family of distributions $p_\theta(\*x\mid\*z)$ with parameters $\theta$. Similar to variational auto-encoders (VAEs)~\cite{kingma2013auto}, VADs are trained by maximizing the marginal distribution $p(\*x)=\int p_\theta(\*x\mid\*z)p(\*z)d\*z$. In practice this integral is expensive or intractable, so the model parameters $\theta$ are learned instead by maximizing the Evidence Lower Bound (ELBO): 
\begin{equation}\label{eq:ELBO}
\begin{aligned}
    \mathcal{V}(\phi,\theta\mid\*x)=&-\text{KL}\big(q_\phi(\*z\mid\*x)\parallel p(\*z)\big)
    +\mathbb{E}_{q_\phi(\*z\mid\*x)}\big[\log p_\theta(\*x\mid\*z)\big],
\end{aligned}
\end{equation}
where $\text{KL}(\cdot\parallel\cdot)$ is the Kullback-Leibler divergence that encourages the posterior distribution to follow the latent prior $p(\*z)$, and $q_\phi(\*z\mid\*x)$ is an approximation of the posterior $p(\*z \mid \*x)$. In VAEs, $q_\phi(\*z\mid\*x)$ is parametrized by a neural network and $\phi$ are the parameters of the encoder. In VADs, $\phi$ are instead learnable similar to the parameters $\theta$ in the decoder $p_\theta(\*x \mid\*z)$. For example, the multivariate Gaussian approximate posterior for a data instance $\*x_i$ is defined as:
\begin{align}
q_\phi(\*z\mid\*x_i) \coloneqq \mathcal{N}(\*z; \*\mu_i, \*\Sigma_i),
\end{align}
where $\phi = \{\*\mu_i, \*\Sigma_i\}$. The reparametrization trick is applied in order to back-propagate the gradients to the mean $\*\mu_i$ and variance $\*\Sigma_i$ in VADs. In comparison, VAEs back-propagate the gradients through the mean $\*\mu_i$ and variance $\*\Sigma_i$ to learn the parameters of the encoder. At inference time, the parameters $\phi$ of the approximate posterior distribution can be estimated by maximizing the ELBO in Eqn.~\ref{eq:ELBO} while the parameters $\theta$ of the decoder are frozen:
\begin{equation}\label{eq:posterior}
    \phi^* = \argmax_{\phi} \ \mathcal{V}(\phi \mid \theta, \*x_i).
\end{equation}

Despite the similarity between VAEs and VADs, prior works~\cite{zadeh2019variational} demonstrate that VADs perform approximate posterior inference more robustly on \emph{incomplete data} and \emph{input domain shifts} than VAEs.

\subsection{Multi-Modal Variational Auto-Decoder}
\label{subsec:MM-VAD}

We consider two modalities $\*x,\*w$ and an \textit{i.i.d.} dataset with paired instances $(\*X,\*W)=\{(\*x_0,\*w_0),\dots,(\*x_N,\*w_N)\}$. We target at learning a joint distribution of both modalities $p(\*x,\*w)$. Like VADs~\cite{zadeh2019variational}, the multi-modal VADs (MM-VADs) are trained by maximizing the ELBO: 
\begin{equation}\label{eq:ELBO2}
    \begin{aligned}
     \mathcal{V}(\phi,\theta\mid\*x,\*w)=&-\text{KL}\big(q_\phi(\*z\mid\*x,\*w)\parallel p(\*z)\big)
     +\mathbb{E}_{q_\phi(\*z\mid\*x,\*w)}\big[\log p_\theta(\*x,\*w\mid\*z)\big],     
    \end{aligned}
\end{equation}
where $\*z$ is the latent variable shared by the two modalities $\*x$ and $\*w$, $p_\theta(\*x,\*w\mid\*z) = p_{\theta_x}(\*x\mid\*z)p_{\theta_w}(\*w\mid\*z)$ under the assumption that the two modalities $\*x$ and $\*w$ are independent conditioned on the latent variable $\*z$ (\ie $\*x\perp\!\!\!\perp\*w\mid\*z$). In practice, $p_{\theta_x}(\*x\mid\*z)$ or $p_{\theta_w}(\*w\mid\*z)$ can be parameterized by different networks for the two modalities $\*x$ and $\*w$ respectively. The parameters $\phi$ of the approximate posterior distribution $q_\phi(\*z\mid\*x,\*w)$ are learnable parameters where $\phi=\{\*\mu,\*\Sigma\}$ under the assumption of multivariate Gaussian posterior distribution. At inference time, the parameters $\phi$ are estimated via maximizing the ELBO with frozen decoder parameters $\theta$:
\begin{equation}\label{eq:posteriror}
    \phi^*=\argmax_{\phi} \ \mathcal{V}(\phi\mid\theta,\*x_i,\*w_i).
\end{equation}

When one of the modalities is missing during inference, the inputs of the missing modalities are simply set to zero. This is the case when we want to infer one modality from the other (\eg 3D reconstruction from 2D sketch). This framework can be trivially extended to learn a joint distribution of more than two modalities.

\subsection{Learning a Joint 2D-3D Prior with MM-VADs}
\label{sec:application}

\begin{figure}[t]
  \begin{minipage}{0.6\textwidth}
    \includegraphics[width=0.7\linewidth]{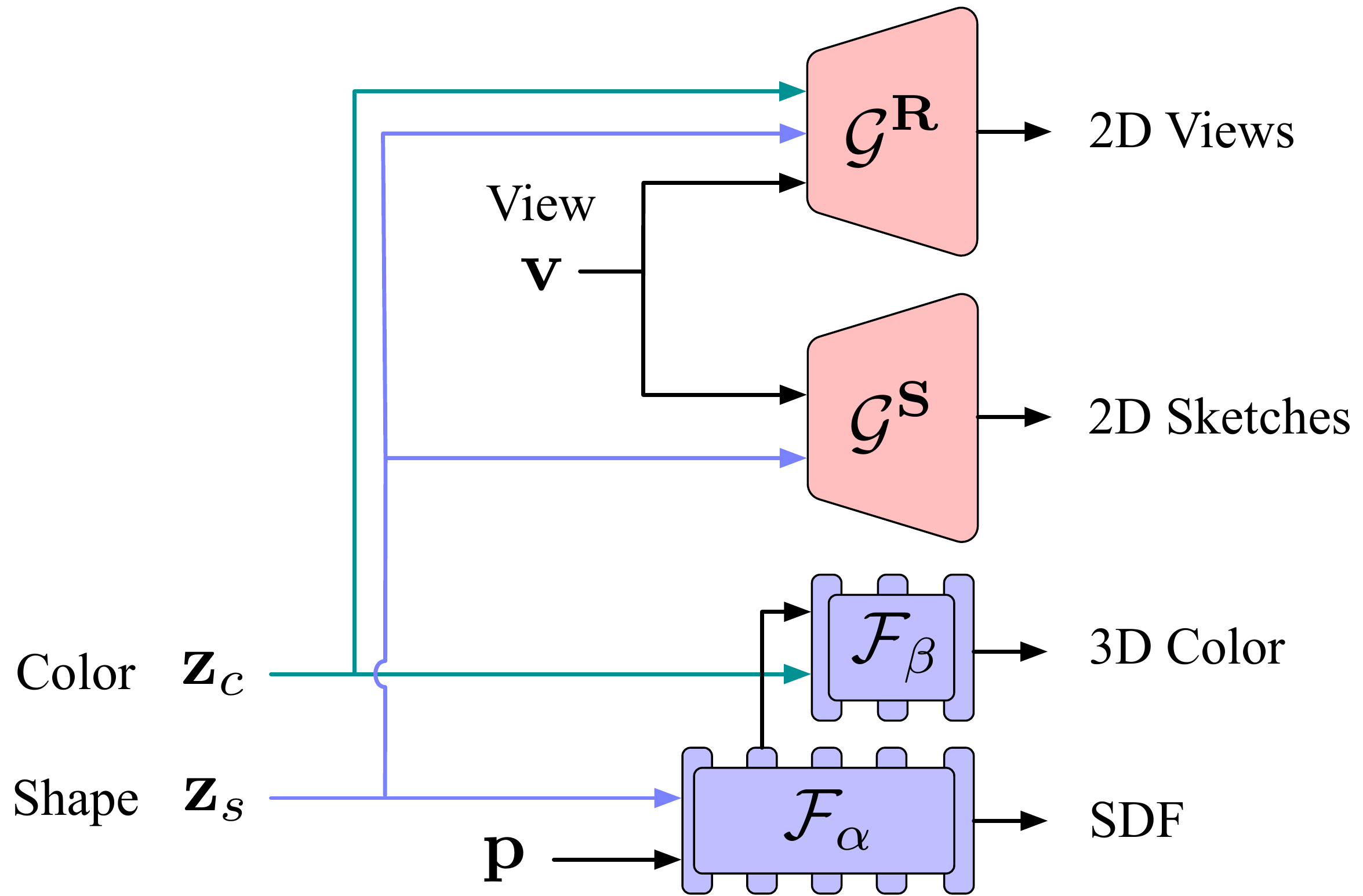}
  \end{minipage}
  \begin{minipage}{0.4\textwidth}
    \caption{\textbf{Network architecture}. We propose a multi-modal variational auto-decoder consisting of compact shape and color latent spaces shared across multiple 2D (\eg sketch, RGB views) or 3D modalities (\eg signed distance function and 3D surface color).}
    \label{fig:network}
  \end{minipage}
\end{figure}

Here we introduce the application of MM-VADs in cross-modal 3D shape generation and manipulation.
Specifically, we learn a joint distribution of 2D and 3D modalities with MM-VADs. Once trained, MM-VADs can be applied to versatile shape generation and editing tasks via a simple posterior inference (or latent optimization).  
We explore three representative modalities, including 3D shape with colorful surface, 2D sketch in grayscale, and 2D rendered image in RGB color, donated as $\*C, \*S, \*R$ respectively. 
Given a dataset $\{(\*C_i,\*S_i,\*R_i)\}$, we target at learning a joint distribution of the three modalities $p(\*C, \*S, \*R)$. 
Fig.~\ref{fig:network} presents the overview of the MM-VADs framework. We provide more details in the following sections.

\noindent\textbf{Joint Latent Space.} The MM-VADs share a common latent space $\mathcal{Z}$ across different modalities (Eqn.~\ref{eq:ELBO2}). Targeting at editing 3D shape and surface color independently, we further disentangle the shared latent space into the shape and color subspaces, denoted as $\mathcal{Z}_s$ and $\mathcal{Z}_c$ respectively. Therefore, each latent code $\*z=\*z_s\oplus\*z_c$, where $\*z_s\in\mathcal{Z}_s$, $\*z_c\in\mathcal{Z}_c$, and $\oplus$ denotes the concatenation operator.

\noindent\textbf{3D Colorful Shape.} Targeting at generating and editing 3D shapes and their appearance, we use the 3D colorful shape as one of our modalities. 
Among various representations of 3D shapes (\eg voxel, mesh, point clouds), the implicit representations~\cite{park2019deepsdf,mescheder2019occupancy,sitzmann2019scene} model 3D shapes as isosurfaces of functions and are capable of capturing high-level details.  
We adopt the DeepSDF~\cite{park2019deepsdf} to regress the signed distance functions (SDFs) from point samples directly using a MLP-based \textit{3D shape network} $\mathcal{F}_\alpha(\*z_s\oplus\*p)$, whose input is a shape latent code $\*z_s\in\mathcal{Z}_s$ and 3D coordinates $\*p \in \mathbb{R}^3$. 
We predict the surface color with another feed-forward \textit{3D color network} $\mathcal{F}_\beta(\*z_c\oplus\*z^k_s)$, whose input is a color latent code $\*z_c\in\mathcal{Z}_c$ and the intermediate features from the $k$-th layer of 3D shape network $\mathcal{F}_\alpha$.
The generator of the 3D modality $\mathcal{G}^{\*C}$ is the combination of the 3D shape and color network:
\begin{equation}
\mathcal{G}^{\*C}(\*z_s\oplus\*z_c\oplus\*p)=\big\{\mathcal{F}_\alpha(\*z_s\oplus\*p),\mathcal{F}_\beta(\*z_c\oplus\*z^k_s)\big\}.
\end{equation}
Both networks are trained using the same set of spatial points. The objective function $\mathcal{L}^{\*C}$ for $\mathcal{G}^{\*C}$ is the $\mathcal{L}_1$ loss defined between the prediction and the ground-truth SDF values and surface colors on the sampled points.

\noindent\textbf{2D Sketch.}
The 2D sketch depicts the 3D structures and provides a natural way for the user to manipulate the 3D shapes.
For the purpose of generalization, we adopt a simple and standard fully convolutional network~\cite{radford2015unsupervised} as our sketch generator $\mathcal{G}^\text{S}(\*z_s\oplus\*v)$ with the shape code $\*z_s\in\mathcal{Z}_s$ and the viewpoint $\*v$ as input. The objective function $\mathcal{L}^{\*S}$ is defined as a cross-entropy loss between the reconstructed and ground-truth sketches. 

\noindent\textbf{2D Rendering.} 
The 2D color rendering reflects a view-dependent appearance of the 3D surface. 
Drawing 2D scribbles on the renderings provides an efficient and straightforward interactive tool for the user to edit the 3D surface color. 
Similar to the 2D sketch modality, we use the standard fully convolutional architecture~\cite{radford2015unsupervised} as our 2D rendering generator $\mathcal{G}^{\*R}(\*z_s\oplus\*z_c\oplus\*v)$, which takes the concatenation of the shape code $\*z_s\in\mathcal{Z}_s$, the color code $\*z_c\in\mathcal{Z}_c$ and the viewpoint $\*v$.
We adopt Laplacian-$\mathcal{L}_1$ loss~\cite{athar2018latent} to train $\mathcal{G}^{\*R}$: 
\begin{equation}
\label{eq:lapl1}
\begin{aligned}
\mathcal{L}^{\*R}(\*z_i\oplus\*v,\*R_i)
=\frac{1}{N}\sum_j^J4^{-j}\big\|\text{L}^j(\mathcal{G}^{\*R}(\*z_i\oplus\*v))-\text{L}^j(\*R_i)\big\|_1,
\end{aligned}
\end{equation}
where $\*z_i$ is the concatenation of the shape and color codes for the target image $\*R_i$, $N$ is the total number of pixels in the image $\*R_i$, $J$ is the total number of levels of the Laplacian pyramid (\eg 3 by default), and $\text{L}^j(x)$ is the $j$-th level in the pyramid of image $x$~\cite{burt1987laplacian}.
This loss encourages sharper output~\cite{athar2018latent} compared to the standard $\mathcal{L}_1$ or MSE loss.

\noindent\textbf{Summary.} The proposed MM-VAD framework for learning the joint distribution of the three modalities can be learned with the following objective: 
\begin{equation}\label{eq:loss}
    \begin{aligned}
     \mathcal{V}(\phi,\theta\mid\*C,\*S,\*R)=-\text{KL}\big(q_\phi(\*z\mid\*C,\*S,\*R)\parallel p(\*z)\big)&\\
     +\ \mathbb{E}_{q_\phi(\*z\mid\*C,\*S,\*R)}\big[\log p_\theta(\*C,\*S,\*R\mid\*z)\big]&,    
    \end{aligned}
\end{equation}
where the first term regularizes the posterior distribution to a latent prior (\eg $\mathcal{N}(\*0,\*I)$), and the second term can be factorized into three components under the assumption that modalities are independent conditioned on the shared latent variable $\*z$:
\begin{equation}\label{eq:3D}
    \begin{aligned}
    \mathbb{E}_{q_\phi(\*z\mid\*C,\*S,\*R)}\big[\log p_\theta(\*C,\*S,\*R\mid\*z)\big]
    =\ &\mathbb{E}_{q_\phi(\*z\mid\*C)}\big[\log p_\theta(\*C\mid\*z)\big]\\
    +\ &\mathbb{E}_{q_\phi(\*z\mid\*C)}\big[\log p_\theta(\*S\mid\*z)\big]\\
    +\ &\mathbb{E}_{q_\phi(\*z\mid\*C)}\big[\log p_\theta(\*R\mid\*z)\big]\\
    =\ &\mathcal{L}^{\*C}+\mathcal{L}^{\*S}+\mathcal{L}^{\*R},
    \end{aligned}
\end{equation}
where each term corresponds to the reconstruction loss per modality as described above. Notice that the 3D shape modality $\*C$ contains all the information in the latent variable $\*z$, therefore $q_\phi(\*z\mid\*C,\*S,\*R)=q_\phi(\*z\mid\*C)$.

\subsection{Cross-Modal Shape Manipulation with MM-VADs}
\label{subsec:manipulation}

Given an initial latent code $\*z_0$ that corresponds to the initial 3D shape $\mathcal{G}^{\*C}(\*z_0)$ and any 2D control $\mathcal{G}^\*M(\*z_0)$ of the 2D modality $\*M \in \{\*S, \*R\}$, 
the shape manipulation is conducted by optimizing within the latent space to get the updated code $\hat{\*z}$ such that $\mathcal{G}(\hat{\*z})^\*M$ matches the 2D edits $\*e^\*M$:
\begin{equation}\label{eq:editing}
\begin{aligned}
    \hat{\*z} = \argmin_{\*z} \ \mathcal{L}_{\text{edit}}(\mathcal{G}^{\*M}(\*z), \*e^{\*M})+ \mathcal{L}_{\text{reg}}(\*z),
\end{aligned}
\end{equation}
where $\mathcal{L}_{\text{edit}}$ could be any loss (\eg $\mathcal{L}_1$ loss) that encourages the 2D modalities $\mathcal{G}(\hat{\*z})^\*M$ to match the 2D edits $\*e^\*M$, and $\mathcal{L}_{\text{reg}}(\*z)$ encourages the latent code to stay in the latent prior of MM-VADs. We apply the regularization loss proposed in DualSDF~\cite{hao2020dualsdf}:
\begin{equation}\label{eq:reg}
\mathcal{L}_{\text{reg}} = \gamma \max(\|\*z\|^2_2, \beta),
\end{equation}
where $\gamma$ and $\beta$ controls the strength of the regularization loss.
The latent optimization is closely related to the posterior inference (Eqn.~\ref{eq:posteriror}) of MM-VADs.

MM-VADs allows free-form edits $e^\*M$. For example, the edits $e^\*M$ could be local modifications on the sketch or sparse color scribbles on 2D renderings. 
This makes the MM-VADs ideally suited for the interactive 3D manipulation tasks. In comparison, the encoder-decoder networks~\cite{guillard2021sketch2mesh} are not robust to the input domain shift (\eg incomplete data~\cite{zadeh2019variational}) and require re-training per type of user interactions (\eg sketch, color scribble).

\subsection{Cross-Modal Shape Generation with MM-VADs}

\noindent\textbf{Single-View Reconstruction.} Given a single input $\*x^\*M$ of the 2D modality $\*M\in\{\*C, \*R\}$, 
the task of single-view cross-modal shape generation is to reconstruct the corresponding 3D shape satisfying the 2D constraint. Without the need of training one model per pair of 2D and 3D modalities~\cite{guillard2021sketch2mesh,tatarchenko2019single} or designing differentiable renderers~\cite{liu2020dist} for each 2D modalities~\cite{guillard2021sketch2mesh}, like shape manipulation (\S\ref{subsec:manipulation}), this task can be tackled via the latent optimization:
\begin{equation}\label{eq:editing}
\begin{aligned}
    \hat{\*z} = \argmin_{\*z} \ \mathcal{L}_{\text{recon}}(\mathcal{G}^\*M(\*z), \*x^\*M)+ \mathcal{L}_{\text{reg}}(\*z),
\end{aligned}
\end{equation}

\noindent\textbf{Partial-View Reconstruction.} The MM-VADs are flexible to reconstruct 3D shapes from partially visible inputs. More interestingly, when the input is ambiguous, it provides diverse 3D reconstructions by performing the latent optimization with different initialization of the latent code $\*z$. This property has practical applications. For example, the MM-VAD could provide multiple 3D shape suggestions interactively while the user is drawing sketches.

\noindent\textbf{Few-Shot Generation.} Given a few 2D images spanning a subspace in the 3D distribution that represents a certain semantic attribute (\eg armchairs, red chairs), the task of few-shot shape generation is to learn a 3D shape generative model that conceptually aligns with the provided 2D images. Given our pre-trained MM-VAD, we tackle this task by steering the latent space with adversarial loss, borrowing the idea from MineGAN~\cite{wang2020minegan}. Specifically, we learn a mapping function $h_\omega(\*z)$ that maps the prior distribution of the latent space $\*z\sim \hat{p}(\*z)$ (\ie $\mathcal{N}(\mathbf{0},\mathbf{I})$) to a new distribution such that samples from the 2D generators $\mathcal{G}^\*M(h_\omega(\*z))$ aligns the target data distribution $\*x \sim \hat{p}(\*x)$ depicted by the provided 2D images. We apply the WGAN-GP loss~\cite{gulrajani2017improved} with frozen generators to learn the mapping function $h_\omega(\*z)$:
\begin{equation}\label{eq:miner}
\begin{aligned}
\min_{\omega}\max_{\mathcal{D}}\ \mathbb{E}_{\*x\sim\hat{p}(\*x)}\big[\mathcal{D}(\*x)\big]
-\ \mathbb{E}_{\*z\sim p(\*z)}\big[\mathcal{D}(\mathcal{G}^\*M(h_\omega(\*z)))\big],
\end{aligned}
\end{equation}
where both the mapping function $h_\omega$ and the discriminator $\mathcal{D}$ are trained from scratch.

\section{Experiments}
\label{sec:experiments}
This section provides qualitative and quantitative results of the proposed MM-VADs in versatile tasks of 3D shape manipulation (\S~\ref{exp:manipulation}) and generation (\S~\ref{exp:generation}). 

\noindent\textbf{Dataset.} We conduct evaluations and comparisons mainly on 3D ShapeNet dataset~\cite{chang2015shapenet}. 
For 3D shapes, We follow DeepSDF~\cite{park2019deepsdf} to sample 3D points and their signed distances to the object surface. 
The points that are far from the surface (\ie with the absolute distance higher than a threshold) are assigned with a pre-defined background color (\eg white) while points surrounding the surface are assigned with the color of the nearest surface point. 
For 2D sketches, we use suggestive contours~\cite{decarlo2003suggestive} to generate the synthetic sketches.
For 2D renderings, we randomize the surface color of 3D shapes per semantic part. 
We use ShapeNet chairs and airplanes with the same training and test splits as DeepSDF~\cite{park2019deepsdf}. 

\noindent\textbf{Implementation Details.} 
We use an $8$-layer MLP as the 3D shape network which outputs SDF 
and a $3$-layer MLP as the 3D color network which predicts RGB. 
We concatenate the features from the $6$-th hidden layer of the 3D shape network with the color code as the input to the 3D color network. 
We train our MM-VADs using Adam~\cite{kingma2014adam}.
We present more implementation details in the Appendix.

\noindent\textbf{Baselines.}  We use the following state-of-the-arts as our baselines:
\begin{itemize}
    \item\textbf{Encoder-Decoder Networks}~\cite{guillard2021sketch2mesh}. This model is trained per task of 3D generation from 2D modalities (sketches or RGB images). We do not use the differentiable rendering proposed in~\cite{guillard2021sketch2mesh} which requires auxiliary information (\eg segmentation mask, depth) and is applicable to MM-VADs.
    \item \textbf{EditNeRF}~\cite{liu2021editing}. This model edits 3D neural radiance field (including shape and color) by updating the neural network weights based on the user's scribbles. We make comparisons with the pre-trained EditNeRF models. 
\end{itemize}

\begin{figure*}[t]
    \centering
    \begin{tabular}{c}
    \includegraphics[width=0.95\linewidth]{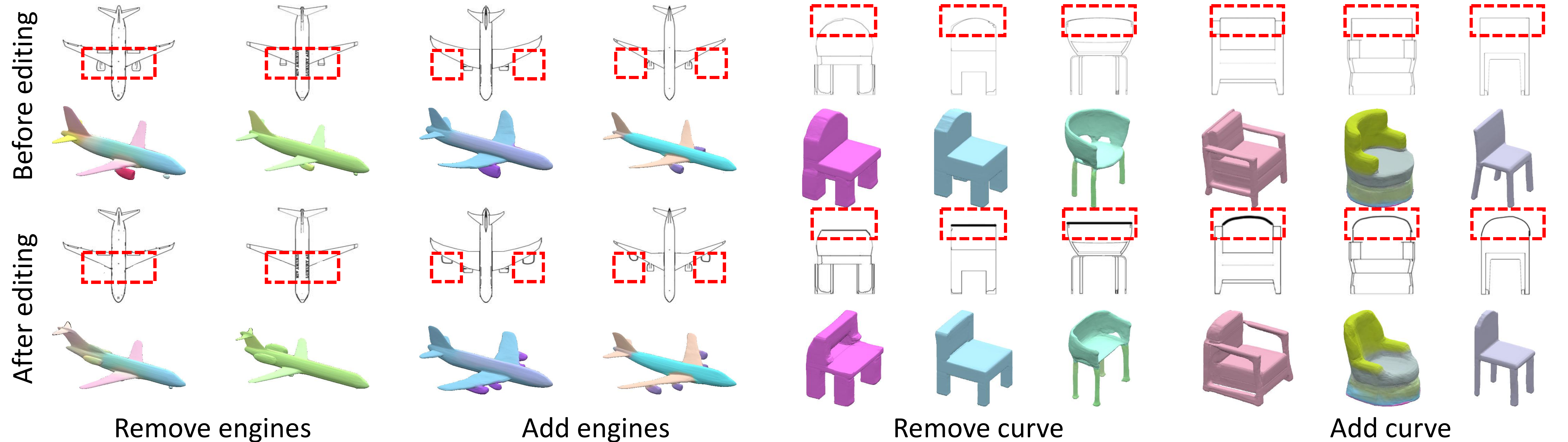} \\
    \end{tabular}
    \caption{\textbf{Editing shape via sketch.} The proposed method enables fine-grained editing of shape geometry, \eg removing the engine of an airplane or reshaping the back of a chair. Interestingly, new engines often appears at the tail of airplane after removing the engines on the wing. This is because airplanes without any engines rarely exist in the domain of our generative model. The edited local regions are highlighted in red bounding boxes.}
    \label{fig:sketch}
\end{figure*}

\subsection{Cross-modal Shape Manipulation}
\label{exp:manipulation}

\setlength{\tabcolsep}{9pt}
\begin{table}[t]
  \caption{\label{tab:sketch}
    \textbf{Editing shape via sketch.} We report the Chamfer distance (CD) between the manually edited shapes and our editing results (\textit{lower} is \textit{better}).}
  \centering
  \begin{tabular}{c c c c c}
\toprule
  & \multicolumn{2}{c}{Airplane} &  \multicolumn{2}{c}{Chair}  \\
  &  $-$ engine & $+$ engine & $-$ curve & $+$ curve\\
\midrule
   Initial shape & 0.096 & \textbf{0.123} & 0.066 & \textbf{0.085}\\
   Edited shape & \textbf{0.059} & 0.134 & \textbf{0.054} & 0.124\\
  \bottomrule
  \end{tabular}
   \label{tab:sketch}
\end{table}

\noindent\textbf{Sketch-Based Shape Manipulation.}
The proposed MM-VADs allow users to edit the fine geometric structures via 2D sketches, as described in \S~\ref{subsec:manipulation}.
We provide users with an interactive interface where users can edit the initial sketch by adding or removing a certain part or even deforming a contour line. 
Fig.~\ref{fig:sketch} presents some qualitative results of sketch-based shape manipulation. 
Interestingly, we find that our manipulation is semantics-aware.
For example, removing the airplane engines on the wings will automatically add new engines to the tail.
Such shape priors are absent in non-generative models (\eg EditNeRF~\cite{liu2021editing}).

It is challenging to quantitatively evaluate the sketch-based shape editing due to the lack of ground-truth paired 3D shapes before and after editing. 
For this reason, prior works~\cite{guillard2021sketch2mesh} report the quantitative results of 3D reconstruction from sketches as a proxy. We follow prior works and report the same quantitative evaluations in Sec.~\ref{exp:generation}. 
Furthermore, we manually edit the 3D shapes presented in Fig.~\ref{fig:sketch} such that their sketches align with the human edits. 
Tab.~\ref{tab:sketch} reports the Chamfer distance (CD) between the manually edited shapes and our editing results. 
We see that CD improves when removing a part, but adding parts unfortunately increases the CD as it induces more changes to the overall shape. This is often desirable,  but the CD metric does not reflect that.

\begin{figure*}[t]
    \centering
    \begin{tabular}{c}
    \includegraphics[width=0.65\linewidth]{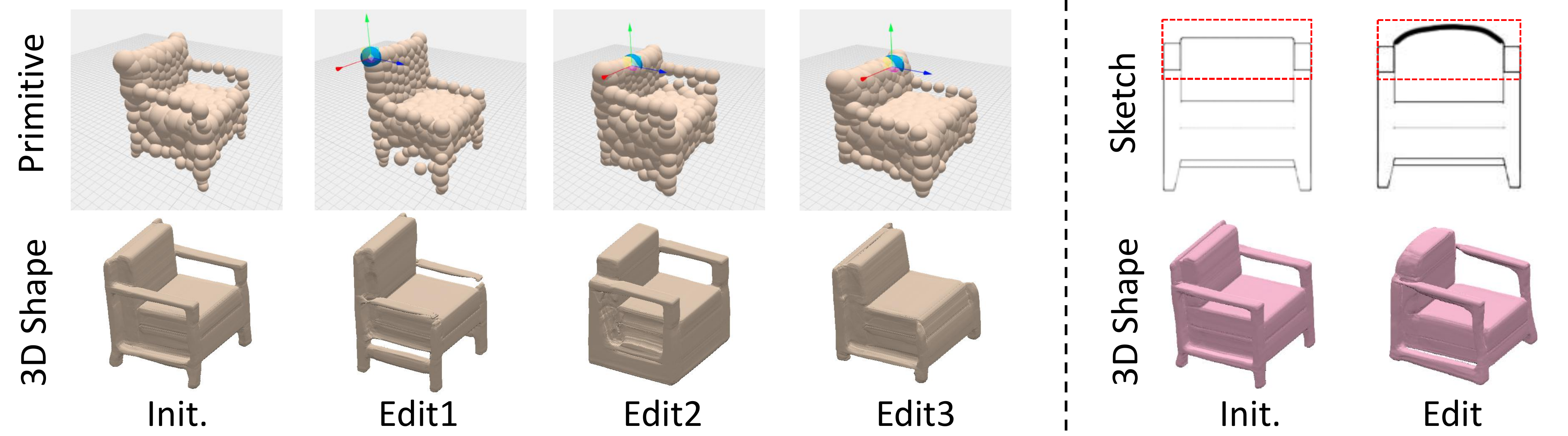} \\
    \end{tabular}
    \caption{\textbf{Comparison with DualSDF}. \textbf{Left:} DualSDF~\cite{hao2020dualsdf} edits 3D shapes via 3D primitives. Editing different primitives on the same part may lead to dramatically different editing results (2nd - 4th columns). \textbf{Right:} our sketch-based interactions is more intuitive for the user.}
    \label{fig:dualsdf}
\end{figure*}

Fig.~\ref{fig:dualsdf} provides a comparison with DualSDF~\cite{hao2020dualsdf}. 
A fair comparison is not possible, as DualSDF edits shapes via 3D primitives instead of 2D views. 
We find that DualSDF requires users to select \textit{right} primitives to achieve certain edits (\eg adding a curve to the chair back). 
In comparison, our sketch-based shape editing is more intuitive.

\begin{figure*} [t]
    \setlength{\tabcolsep}{4pt}
    \newcommand\wifig{0.07\linewidth}
    \centering
     \begin{tabular}{cccccc | ccccc}
     \rotatebox[origin=c]{90}{2D} &
    \includegraphics[width=\wifig, valign=c]{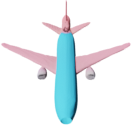} & 
    \includegraphics[width=\wifig, valign=c]{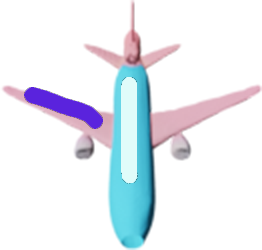} & 
    \includegraphics[width=\wifig, valign=c]{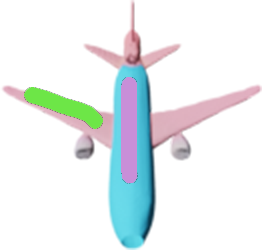} & 
    \includegraphics[width=\wifig, valign=c]{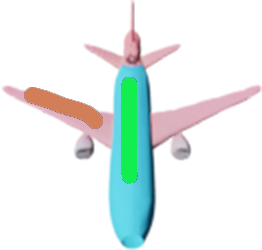} & 
    \includegraphics[width=\wifig, valign=c]{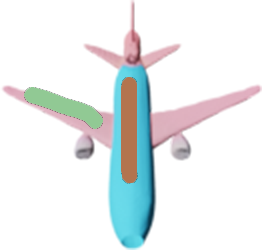} & 
    \includegraphics[width=\wifig, valign=c]{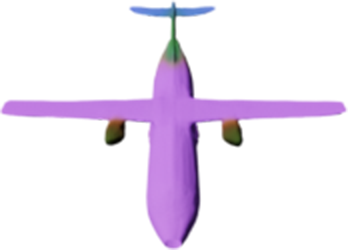} & 
    \includegraphics[width=\wifig, valign=c]{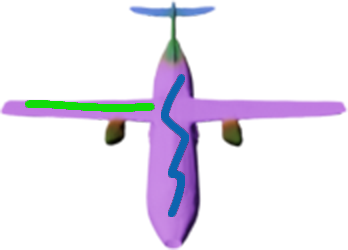} & 
    \includegraphics[width=\wifig, valign=c]{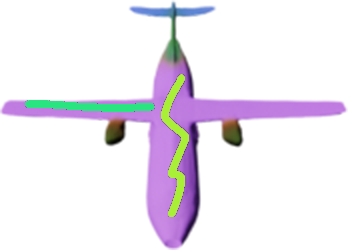} & 
    \includegraphics[width=\wifig, valign=c]{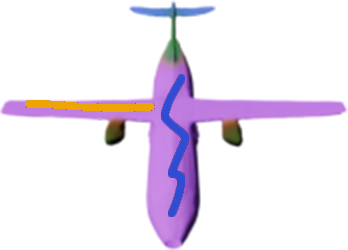} & 
    \includegraphics[width=\wifig, valign=c]{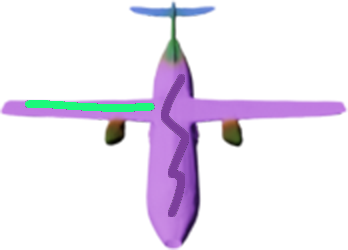} \\
    \rotatebox[origin=c]{90}{3D} &
      \includegraphics[width=\wifig, valign=c]{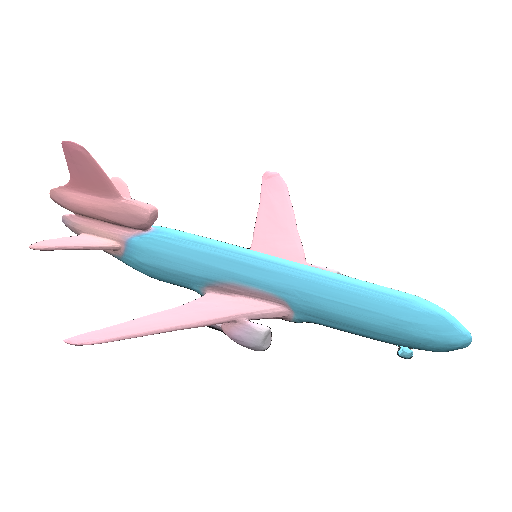} & 
    \includegraphics[width=\wifig, valign=c]{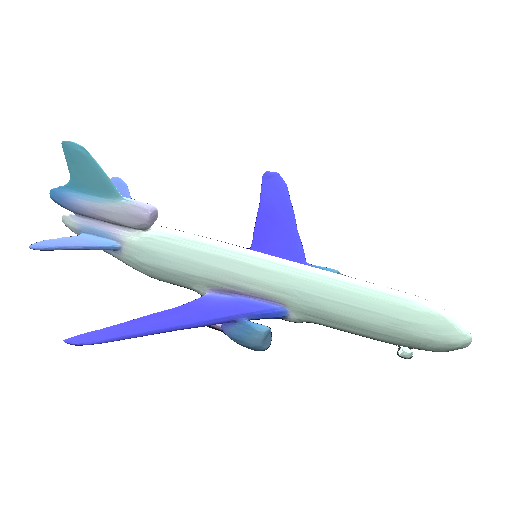} & 
    \includegraphics[width=\wifig, valign=c]{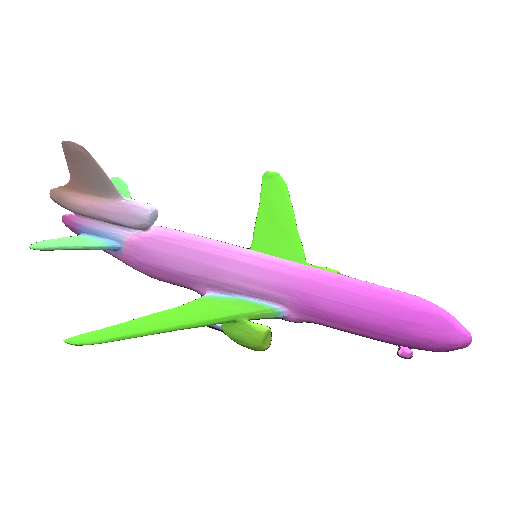} & 
    \includegraphics[width=\wifig, valign=c]{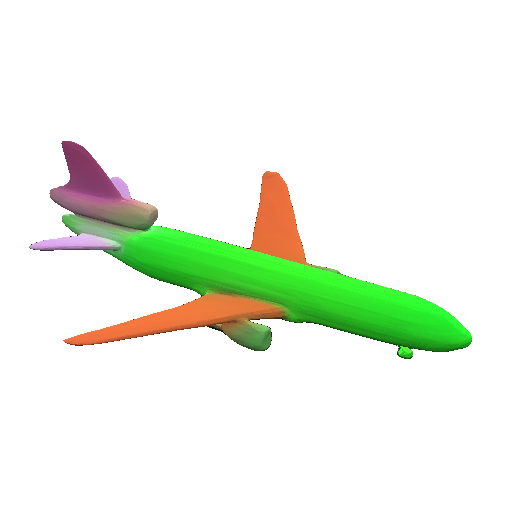} & 
    \includegraphics[width=\wifig, valign=c]{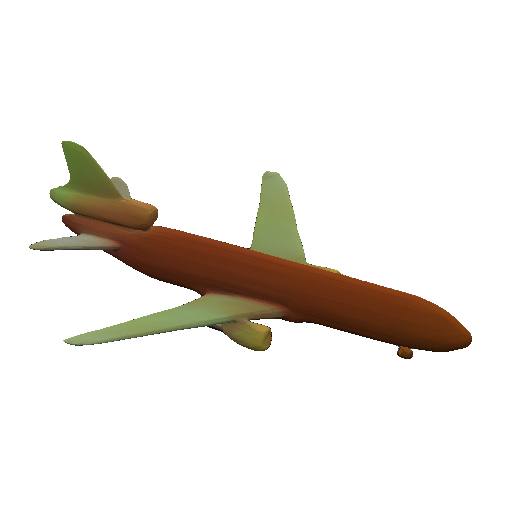} & 
    \includegraphics[width=\wifig, valign=c]{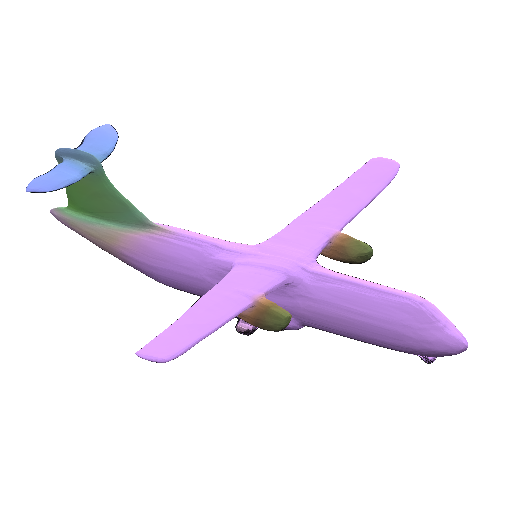} & 
    \includegraphics[width=\wifig, valign=c]{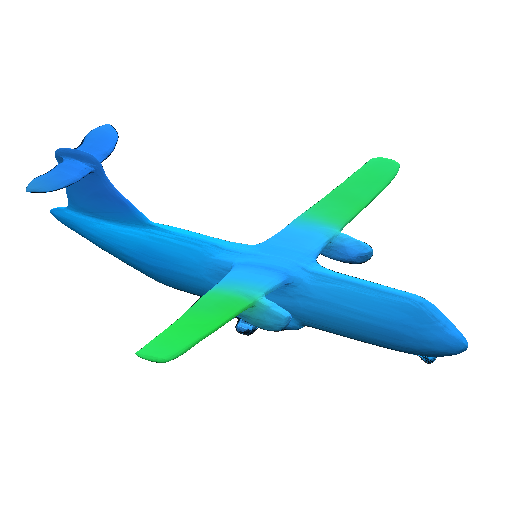} & 
    \includegraphics[width=\wifig, valign=c]{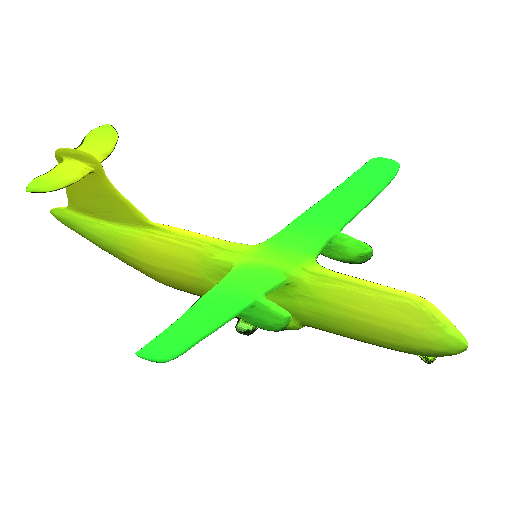} & 
    \includegraphics[width=\wifig, valign=c]{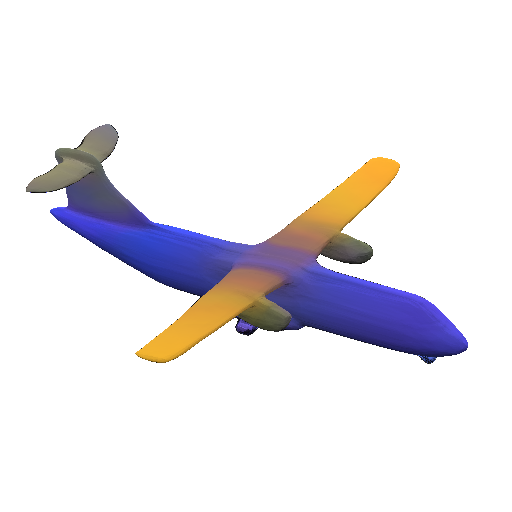} &    \includegraphics[width=\wifig, valign=c]{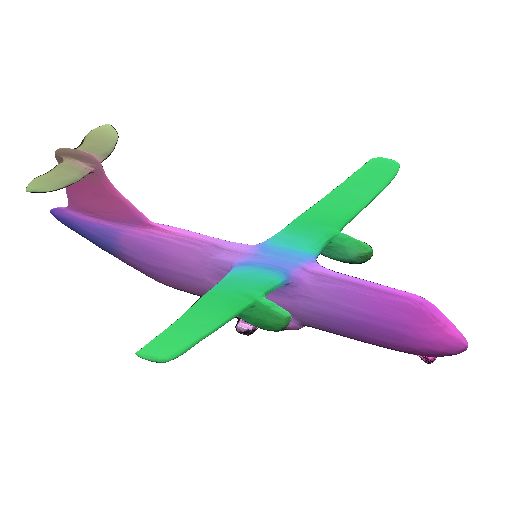} \\
    \rotatebox[origin=c]{90}{2D} &
    \includegraphics[width=\wifig, valign=c]{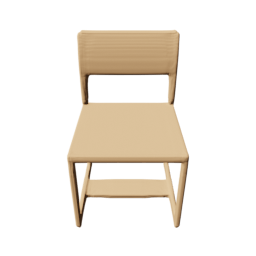} & 
    \includegraphics[width=\wifig, valign=c]{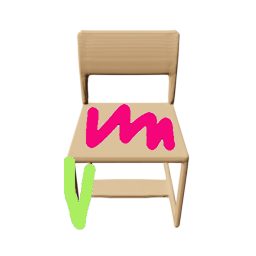} &
    \includegraphics[width=\wifig, valign=c]{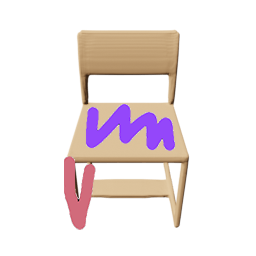} &
    \includegraphics[width=\wifig, valign=c]{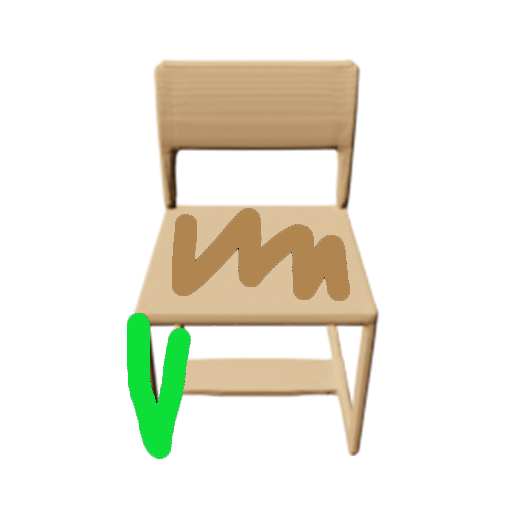} &
    \includegraphics[width=\wifig, valign=c]{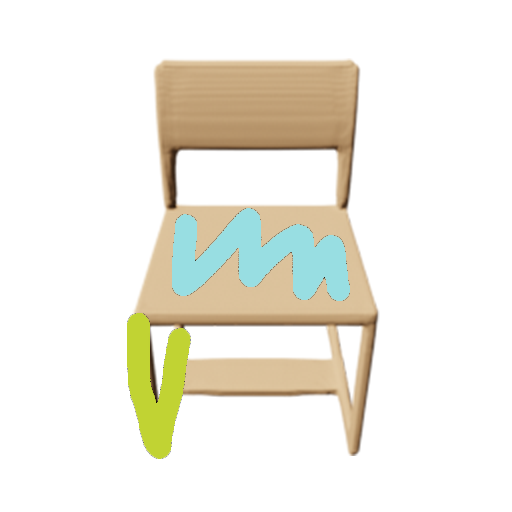} &
    \includegraphics[width=\wifig, valign=c]{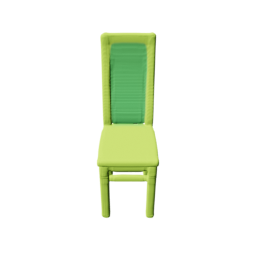} & \includegraphics[width=\wifig, valign=c]{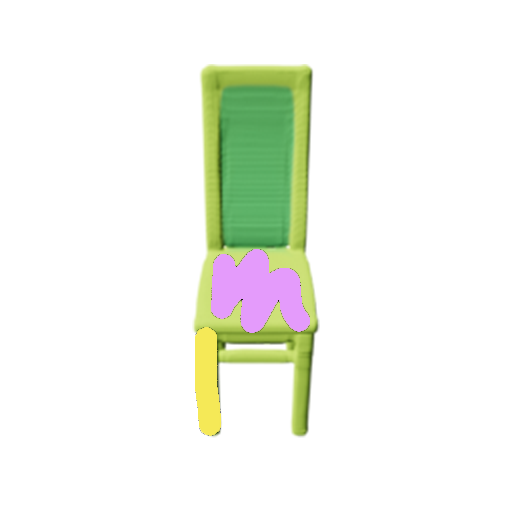}&
    \includegraphics[width=\wifig, valign=c]{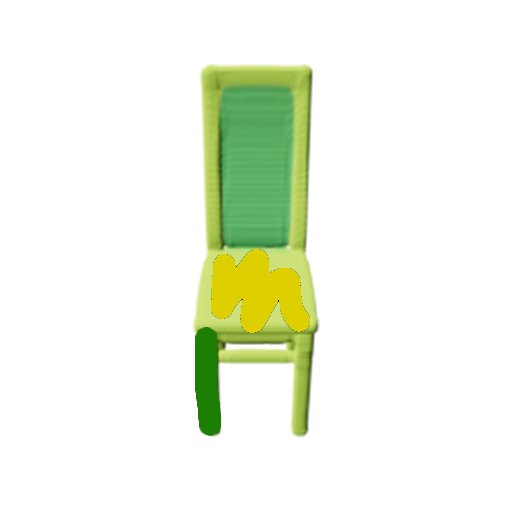} &
    \includegraphics[width=\wifig, valign=c]{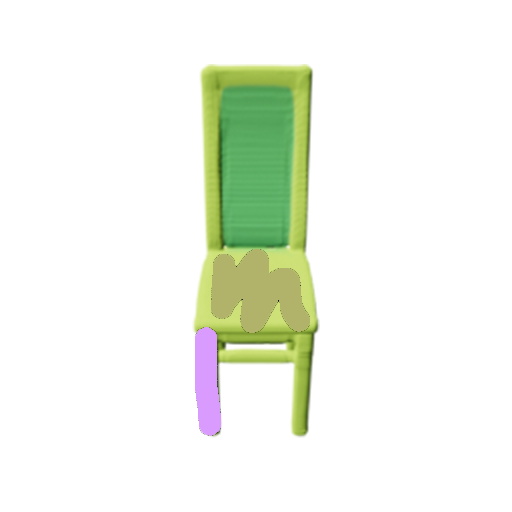} &
    \includegraphics[width=\wifig, valign=c]{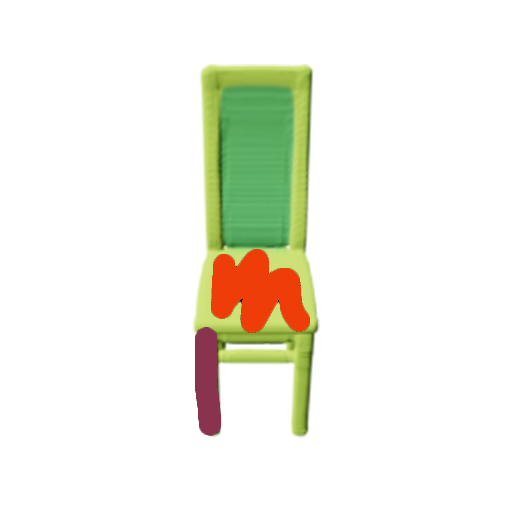} \\
    \rotatebox[origin=c]{90}{3D} &
     \includegraphics[width=\wifig, valign=c]{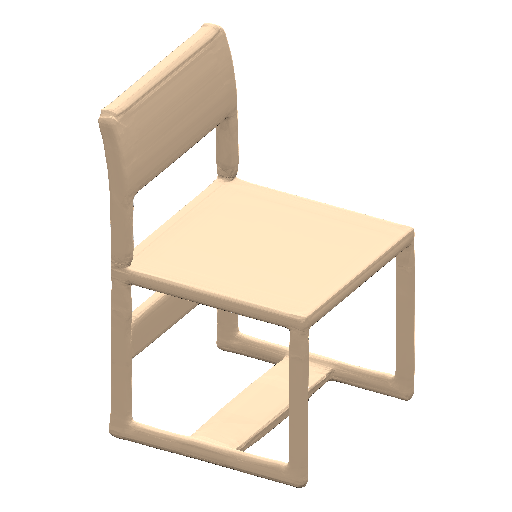}& 
    \includegraphics[width=\wifig, valign=c]{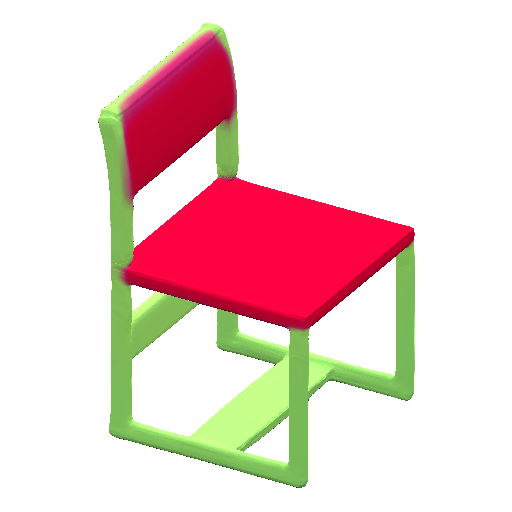} &
    \includegraphics[width=\wifig, valign=c]{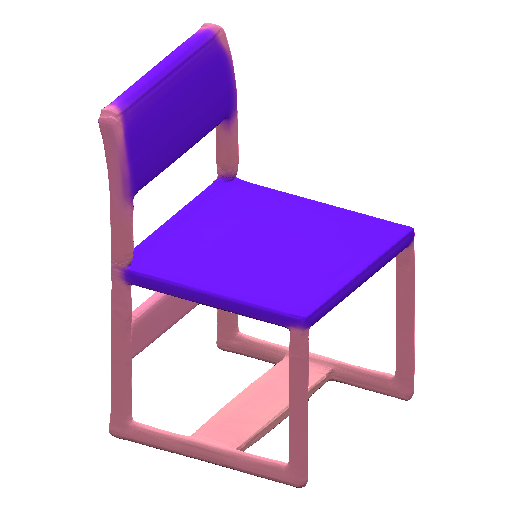} &
    \includegraphics[width=\wifig, valign=c]{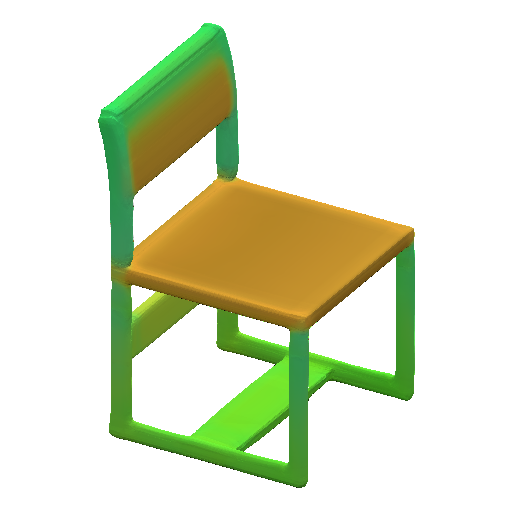} &
    \includegraphics[width=\wifig, valign=c]{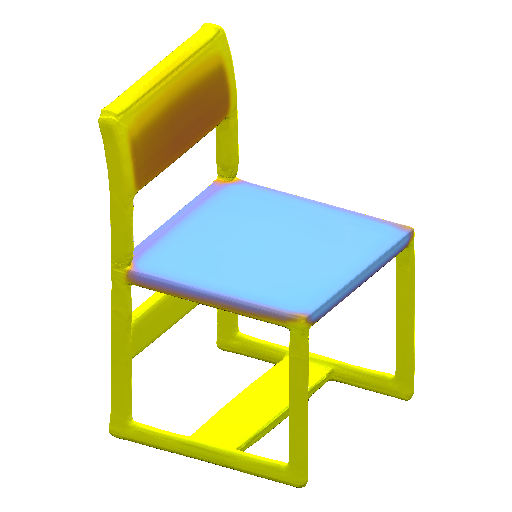} &
    \includegraphics[width=\wifig, valign=c]{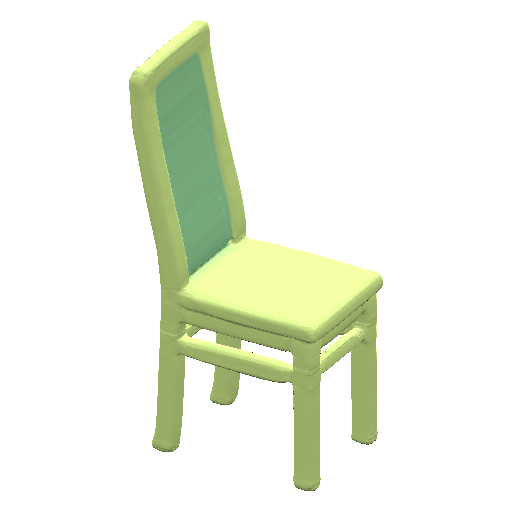} & \includegraphics[width=\wifig, valign=c]{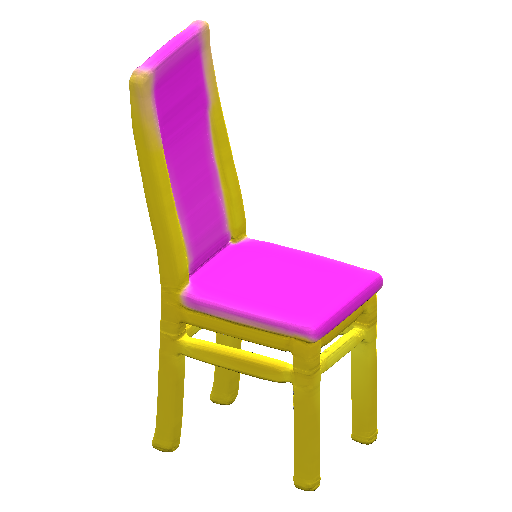}&
    \includegraphics[width=\wifig, valign=c]{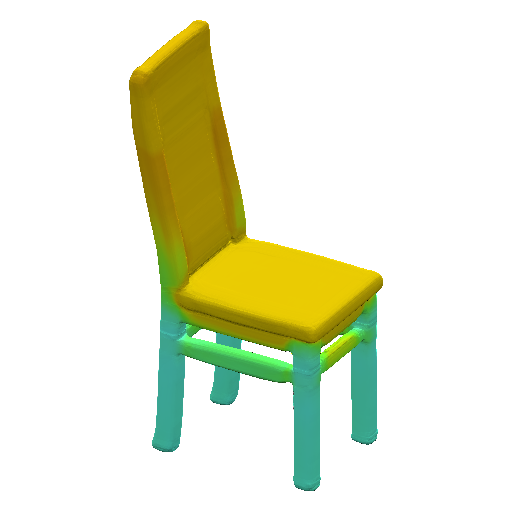} &
    \includegraphics[width=\wifig, valign=c]{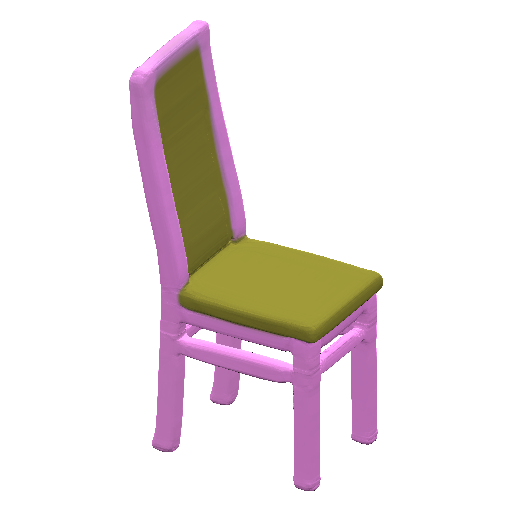} &
    \includegraphics[width=\wifig, valign=c]{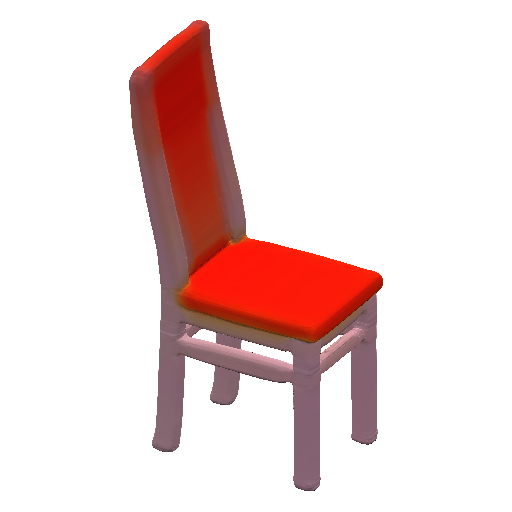} \\
    & (a) Init. &\multicolumn{4}{c}{(b) Color editing} & (a) Init. &\multicolumn{4}{c}{(b) Color editing} \\
    \end{tabular}
    \caption{\textbf{Editing shape via color scribble.} \textbf{(a)} presents the initial 2D and 3D view of the object. \textbf{(b)} shows the 2D color scribbles and 3D color editing results.}
    \label{fig:scribble}
\end{figure*}

\noindent\textbf{Scribble-Based Color Manipulation.}
MM-VADs allow users to edit the appearance of 3D shapes via color scribbles. 
Fig.~\ref{fig:scribble} shows that MM-VADs propagate the sparse color scribbles into desired regions (\eg from the left wing of the airplanes to the right, from the left leg of chairs to the right).
We provide more color editing results with diverse color scribbles in the appendix.
As a quantitative evaluation ,we select 10 shapes per category (including airplanes and chairs) and edit the surface color to make it visually similar to reference shapes with same geometry yet different surface color. 
The editing quality is measured by the similarity between the renderings of the edited 3D shapes and the reference shapes.
Tab.~\ref{tab:scribble} reports the PSNR and LPIPS~\cite{zhang2018perceptual} metrics of the evaluation. 
The surface color of 3D shapes is much closer to the reference after editing, compared to the initial shapes, suggesting the effectiveness of our MM-VAD model in editing color via scribbles. 

A similar task has recently been explored in EditNeRF~\cite{liu2021editing}.
However, an apple-to-apple comparison with EditNeRF is not possible due to the intrinsically different 3D representations (NeRF~\cite{mildenhall2020nerf} vs SDFs~\cite{park2019deepsdf}).
Moreover, the proposed MM-VADs are generative models while EditNeRF is non-generative; 
The MM-VADs bridge 2D and 3D via shared latent spaces while EditNeRF relies on differentiable rendering. 
We present more detailed comparisons in the appendix.
We provide qualitative comparisons with EditNeRF on chairs with similar structures using their pre-trained models.  
Fig.~\ref{fig:editnerf} shows that the color editing from MM-VADs is on par with EditNeRF. 
The MM-VADs achieve the editing via simple latent optimization (Eqn.~\ref{eq:editing}), while EditNeRF requires updating the network weights per instance and fails to generate meaningful color editing results via optimizing the color code alone. 
Furthermore, MM-VADs take $0.06$ seconds per edit and $6.78$ seconds to render our 3D shapes into $256\times256$ RGB images, while EditNeRF takes over a minute per edit including rendering.

\setlength{\tabcolsep}{9pt}
\begin{table}[ht!]
  \caption{\label{tab:scribble}
    \textbf{Quantitative results of editing 3D via 2D scribbles.} We edit the surface color of 3D shape based on reference shapes, and report the similarity between the editing results and the target (bottom row). As a reference, we also report the metrics before editing (top row).}
  \centering
  \begin{tabular}{c c c c c}
    \toprule
      \multirow{2}{*}{Methods} & \multicolumn{2}{c}{Airplane} &  \multicolumn{2}{c}{Chair}  \\
  & PSNR $\uparrow$ & LPIPS $\downarrow$ & PSNR $\uparrow$ & LPIPS $\downarrow$\\
  \midrule
    Initial & 19.84 & 0.23 & 16.20 & 0.33\\
    Edited  & \textbf{26.41} & \textbf{0.13} & \textbf{22.08} & \textbf{0.20}\\
    \bottomrule
  \end{tabular}
\end{table}

\begin{figure} [t]
    \setlength{\tabcolsep}{2pt}
    \newcommand\wifig{0.08\linewidth}
    \centering
     \begin{tabular}{ccccccc}
      \rotatebox[origin=c]{90}{\cite{liu2021editing}} &
    \includegraphics[width=\wifig, valign=c]{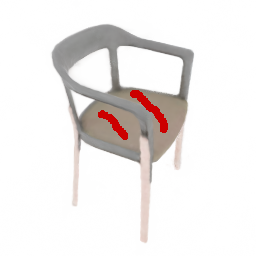} &
    \includegraphics[width=\wifig, valign=c]{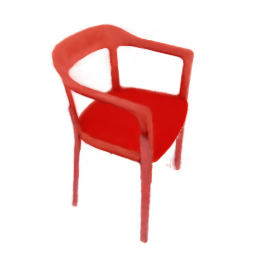} &
    \includegraphics[width=\wifig, valign=c]{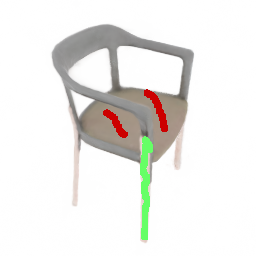} &
    \includegraphics[width=\wifig, valign=c]{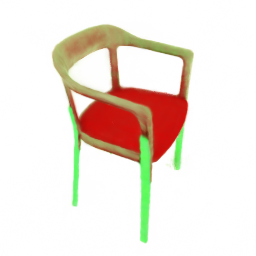} &
    \includegraphics[width=\wifig, valign=c]{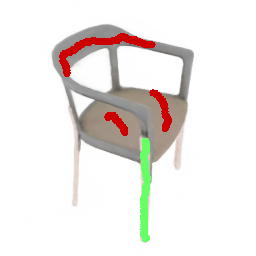} &
    \includegraphics[width=\wifig, valign=c]{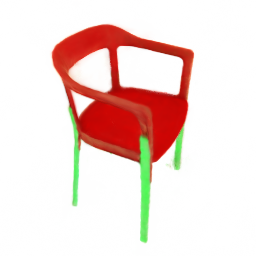} \\
    \rotatebox[origin=c]{90}{Ours} &
    \includegraphics[width=\wifig, valign=c]{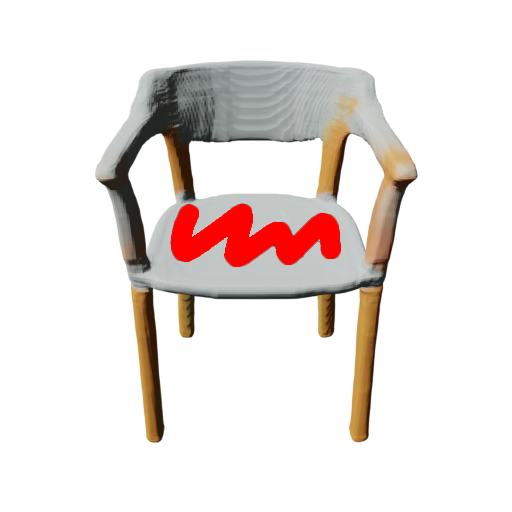} &
    \includegraphics[width=\wifig, valign=c]{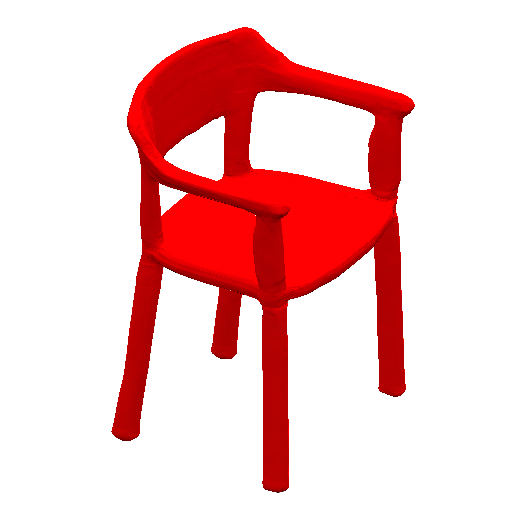} &
    \includegraphics[width=\wifig, valign=c]{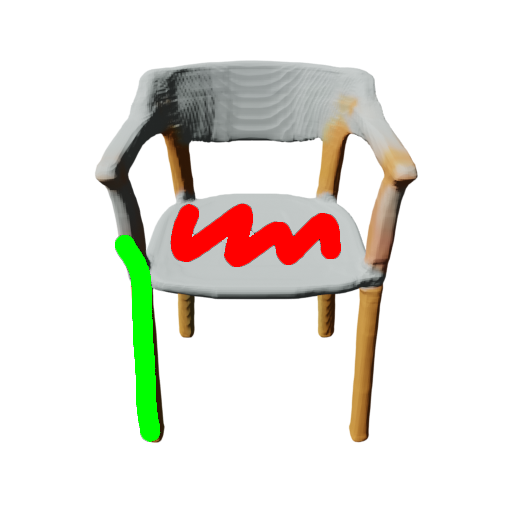} &
    \includegraphics[width=\wifig, valign=c]{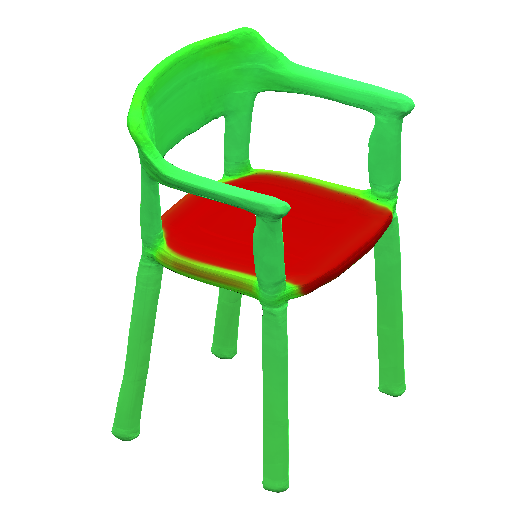} &
    \includegraphics[width=\wifig, valign=c]{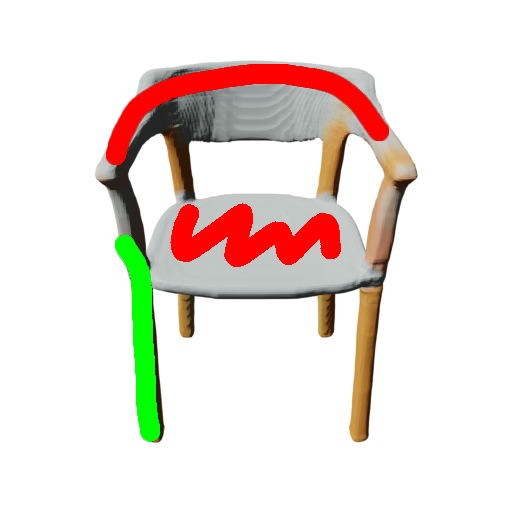} &
    \includegraphics[width=\wifig, valign=c]{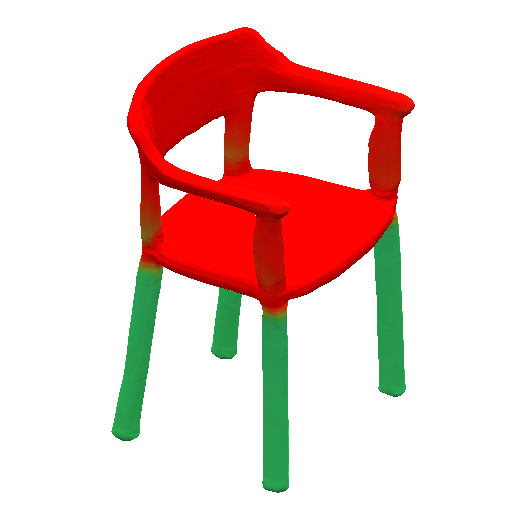} \\
    \end{tabular}
    \caption{\textbf{Comparison with EditNeRF}. Our model (\textbf{bottom}) achieves comparable editing performance with EditNeRF~\cite{liu2021editing} (\textbf{top}). We provide three color edits on 2D views (\textbf{odd columns}), each followed by the 3D editing result (\textbf{even columns}).}
    \label{fig:editnerf}
\end{figure}

\subsection{Cross-Modal Shape Generation}
\label{exp:generation}

\begin{figure*}[t]
    \centering
    \begin{tabular}{c}
    \includegraphics[width=0.9\linewidth]{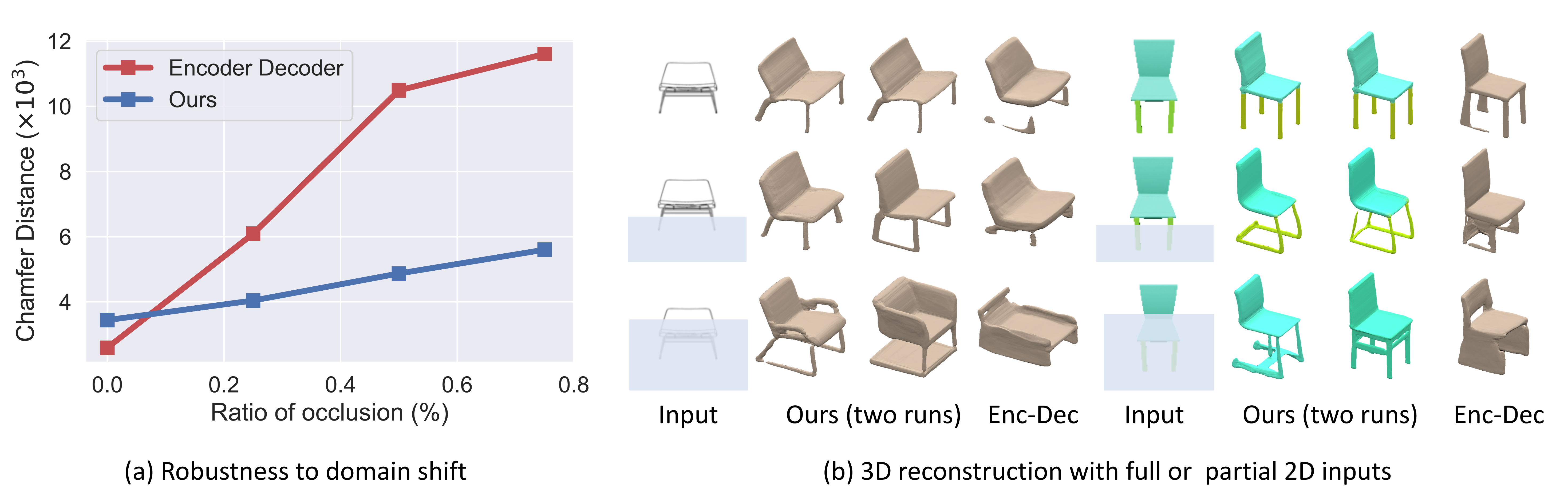} \\
    \end{tabular}
    \caption{\textbf{(a) Robustness to domain shift.} We report the Chamfer distance (\emph{lower} is \emph{better}) between 3D reconstructions and the groundtruth under different ratios of image occlusion. \textbf{(b) 3D reconstruction with full or partial 2D inputs.} When the full views are available, our model produces consistent 3D reconstruction in different trials. When only partial views are given, our model produces multiple different 3D reconstructions. In comparison, the encoder-decoder networks~\cite{guillard2021sketch2mesh} trained on full-view sketches are not robust to the domain shift induced by occlusion and unable to provide multiple 3D shapes given partial views. Notice that the predictions of surface color is not available in the encoder-decoder networks from the prior work~\cite{guillard2021sketch2mesh}.}
    \label{fig:curves}
\end{figure*}

\noindent\textbf{Single-View and Partial-View Shape Reconstruction.}
Fig.~\ref{fig:curves} compares the performance of our model and the encoder-decoder networks~\cite{guillard2021sketch2mesh} under different occlusion ratios in the lower part of the objects in 2D views. 
The proposed model only has a slight performance drop as the occluded parts increase (Fig.~\ref{fig:curves}a), mainly because of the ambiguity of 3D reconstruction given partial views. 
In fact, our reconstructions results fit the partial views quite well.
Even though our model performs slightly worse than the encoder-decoder networks on full-view inputs, the proposed model is more robust to the input domain shift. This is because compared to task-specific training, our model achieves a better trade-off between reconstruction accuracy and domain generalization.
More interestingly, our model can achieve diverse and reasonable 3D reconstruction by sampling different initialization for latent optimization (Fig.~\ref{fig:curves}b).

\begin{figure*}[t]
    \setlength{\tabcolsep}{0.6pt}
    \newcommand\wifig{\linewidth}
    \centering
     \begin{tabular}{cccc}
     \includegraphics[width=0.9\wifig, valign=c]{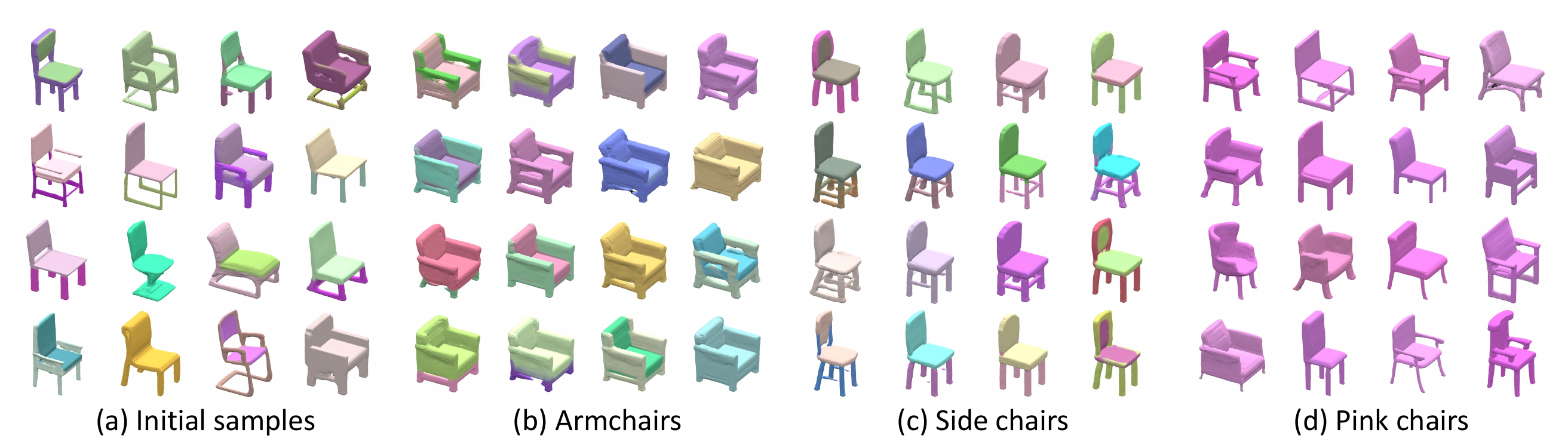} 
    \end{tabular}
    \caption{\textbf{Few-shot cross-modal shape generation.} (a) presents random 3D samples from our model before the adaptation. Given a few 2D exemplars of a certain category (\eg armchair), our model can be adapted to generate corresponding 3D shapes (b-d).}
    \label{fig:few-shot}
\end{figure*}

\noindent\textbf{Few-Shot Shape Generation.} 
The proposed method is able to adapt the pre-trained multi-modal generative model with as few as $10$ training samples of a specific 2D modality. Fig.~\ref{fig:few-shot} presents some of the few-shot cross-modal shape generation results. 
To quantitatively evaluate the few-shot shape generation performance, we render the 3D shapes into 2D RGB images and report the Frechet Inception Distance (FID) scores~\cite{heusel2017gans} between the rendered images and the ground-truth samples. 
Since the FID score is not sensitive to the semantic difference between two image sets, we also report the classification error on the random samples from the model before and after the adaptation. 
Specifically, we train a binary image classifier to identify the target image categories (\eg armchairs vs. other chairs), and we run the trained classifier on the 2D renderings of the 3D samples before and after the adaptation. 
As presented in Tab.~\ref{tab:few-shot}, our pre-trained generative model can be effectively adapted to a certain shape subspace given as few as $10$ 2D examples.
This capability allows us to agilely adapt our generative model to a subspace defined by a few unlabelled samples, so that users can easily narrow down the target shape during the manipulation by providing a few samples of a common attribute, such as a specific category, style, or color.
We are unaware of any prior works that can tackle this task in the literature.
The 2D examples used to adapt the pre-trained generative model are provided in our appendix.

\setlength{\tabcolsep}{9pt}
\begin{table}[t]
    \caption{\textbf{Quantitative results of few-shot cross-modal shape generation.} We report Frechet Inception Distance (FID)  (\emph{lower} is \emph{better}) and classification error (Cls. Err)  (\emph{lower} is \emph{better}). We effectively adapt the pretrained multi-modal VAD model using a few 2D images to a desired 3D shape generator. As a reference, we report the metrics before the few-shot adaptation (top row).}
    \label{tab:few-shot}
   \centering
   \begin{tabular}{c r cccc}
    \toprule
  Stage & Metrics  & Arm & Side & Red & Avg.\\
   \midrule
\multirow{2}{*}{Init.} & FID $\downarrow$ & 138.1 & 95.2  &  93.7  & 109.0  \\
 & Cls.Err. $\downarrow$ & 0.79 & 0.64 & 0.82 & 0.75 \\
 \midrule
\multirow{2}{*}{Adapt.} & FID $\downarrow$ &\textbf{130.4}  & \textbf{92.4} & \textbf{93.0} &   \textbf{105.3} \\ 
 & Cls.Err. $\downarrow$ & \textbf{0.01} & \textbf{0.10} & \textbf{0.00} & \textbf{0.04} \\ 
    \bottomrule
  \end{tabular}
\end{table}

\begin{figure}[t]
  \begin{minipage}[c]{0.5\textwidth}
        \begin{tabular}{c}
    \includegraphics[width=\linewidth]{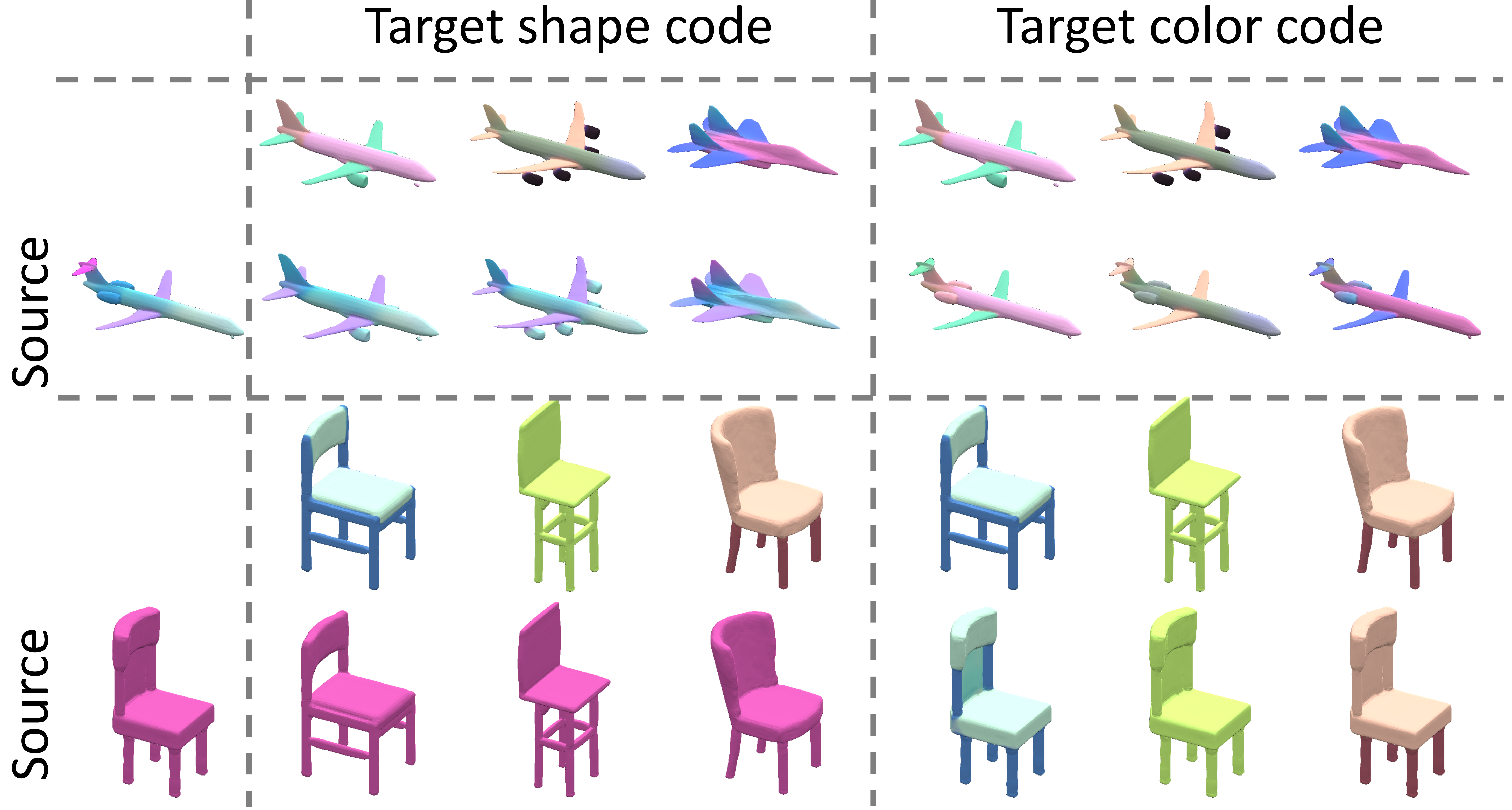} \\
    \end{tabular}
    \caption{\textbf{Shape and color transfer.} The reference 3D shapes (top row) provide the shape codes or color codes for each source instances (first column).}
     \label{fig:transfer}
  \end{minipage}\hfill
  \begin{minipage}[c]{0.45\textwidth}
    \centering
     \begin{tabular}{cc}
    \includegraphics[width=0.2\linewidth, valign=c]{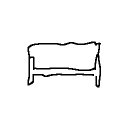} & 
    \includegraphics[width=0.2\linewidth, valign=c]{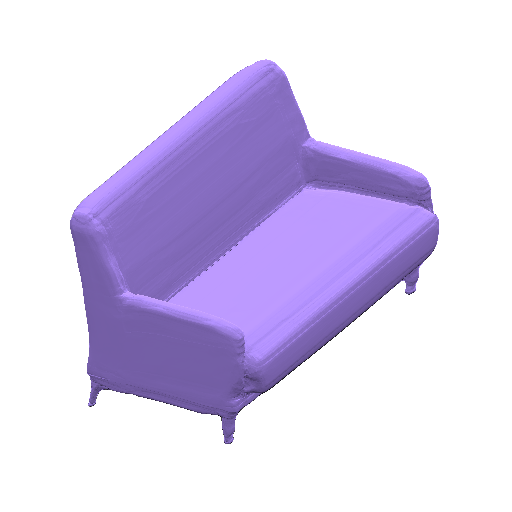} \\
    (a) Real sketch & (b) 3D Recon. \\ 
    \includegraphics[width=0.2\linewidth, valign=c]{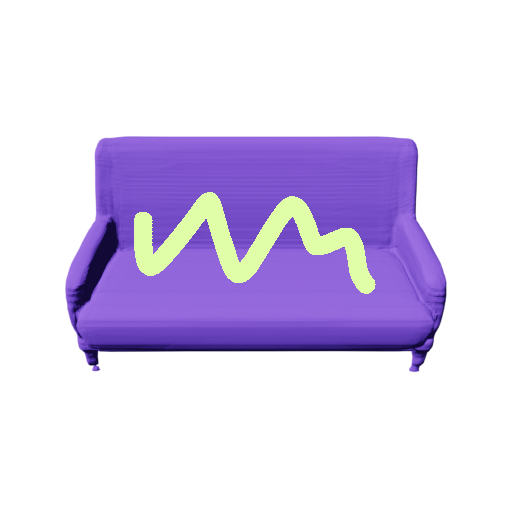} &
    \includegraphics[width=0.2\linewidth, valign=c]{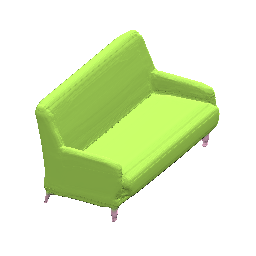}\\
    (c) Edit color & (d) Result \\
    \end{tabular}
    \caption{Our model enables consecutive 3D reconstruction and manipulation given a hand-drawn sketch.}
    \label{fig:real}
  \end{minipage}
\end{figure}

\noindent\textbf{Shape and Color Transfer.} Transferring shape and color across different 3D instances can be achieved by simply swapping the latent codes. Fig.~\ref{fig:transfer} shows that the shape and color are well disentangled in the proposed generative model. The transfer results also are semantically meaningful, \ie the color is only transferred across the same semantic parts (\eg seats for the chair, wings for the airplane) even though the geometry of the source and target instances are quite different.

\subsection{Case Study on Real Images} 
The workflow of 3D designers usually starts by drawing a 2D sketch to portray the coarse 3D geometry and then colorizes the sketch to depict the 3D appearance. 
These 2D arts are used as a reference to build 3D objects.
Undoubtedly this procedure requires extensive human efforts and expertise. 
Such tasks can be automated with our MM-VADs. 
As shown in Fig.~\ref{fig:real}, we first reconstruct the 3D shape from a hand-drawn sketch. We then assign a surface color by randomly sampling a color code from the latent space of the MM-VADs, which can be easily edited by drawing color scribbles on the surface. 
Our model does not require any re-training on each of these steps and provides a tool to conduct shape generation and color editing consecutively. 
Such a task is infeasible with the existing works that train an encoder-decoder network to predict 3D shape from sketch~\cite{guillard2021sketch2mesh}.

\section{Discussion}
\label{sec:conclusion}

We propose a multi-modal generative model which bridges multiple 2D and 3D modalities through a shared latent space.
One limitation of the proposed method is that we are only able to provide editing results in the prior distribution of our generative model (see appendix for more details). 
Despite this limitation, our model has enabled versatile cross-modal 3D generation and manipulation tasks without the need of re-training per task and demonstrates strong robustness to input domain shift. 

\paragraph{Acknowledgements.} Subhransu Maji acknowledges support from NSF grants \#1749833 and \#1908669. Our experiments were partially performed on the University of Massachusetts GPU cluster funded by the Mass. Technology Collaborative.

\clearpage
\appendix
\noindent{\Large \textbf{Appendix}}

\section{Implementation Details}\label{sec:implementation}

The implementation details of 3D shape and color networks are included in the main text. Here we provide additional implementation details. 
\begin{itemize}
    \item \textbf{Joint latent space.} The shape and color latent codes are both of dimension $128$ throughout our experiments. We observe that lower-dimensional latent codes (\eg $32$) lead to worse shape reconstruction. 
    \item \textbf{2D sketch and renderings.} The image resolution of all 2D modalities is set to $128\times128$. We use the generator architecture from DCGAN~\cite{radford2015unsupervised} for all 2D modalities. 
    \item \textbf{Few-shot shape generation.} We use the discriminator from DCGAN~\cite{radford2015unsupervised}. The mapping function $h_w(\*z)$ in the MineGAN framework~\cite{wang2020minegan} is a two-layer MLP with batch normalization~\cite{ioffe2015batch} and a ReLU activation function.
    \item  \textbf{Latent optimization.} In the task of shape and appearance manipulation, we conduct the latent optimization for $5$ steps starting from a known initial latent code that corresponds to the initial 2D and 3D instances. The hyperparameter $\gamma$ and $\beta$ in Eqn.11 is 0.02 and 0.5 respectively by default. 
    For the single-view shape generation tasks, we run multiple trials of latent optimization from different randomly sampled latent codes. 
    The optimized code with minimal reconstruction loss is used as the final result. 
    We observe that such multi-trial optimization significantly stabilizes the performance of 3D reconstruction (see Sec.~\ref{sec:ablation} for more details).
\end{itemize}

\section{Ablation study}\label{sec:ablation}

The latent optimization is crucial to the performance of our shape reconstruction and manipulation tasks. In this section, we provide ablation studies on the regularization loss (Eqn.~10 in the main text) and the multi-trial latent optimization method (as described in Sec.~\ref{sec:implementation}). 

\begin{figure} [h!]
    \setlength{\tabcolsep}{0.7pt}
    \newcommand\wifig{0.2\linewidth}
    \centering
     \begin{tabular}{ccc}
    \includegraphics[width=\wifig, valign=c]{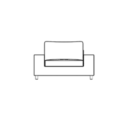} & 
    \includegraphics[width=\wifig, valign=c]{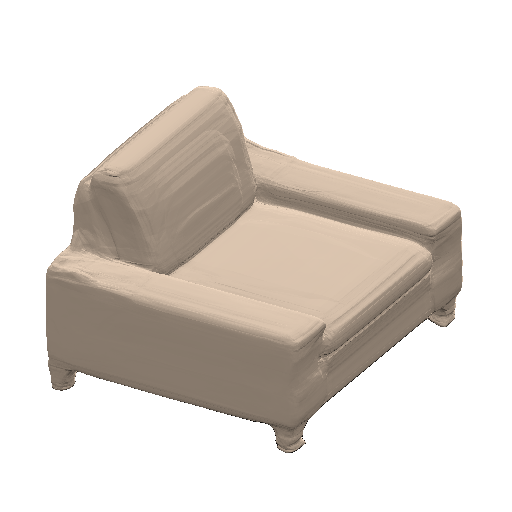} &
    \includegraphics[width=\wifig, valign=c]{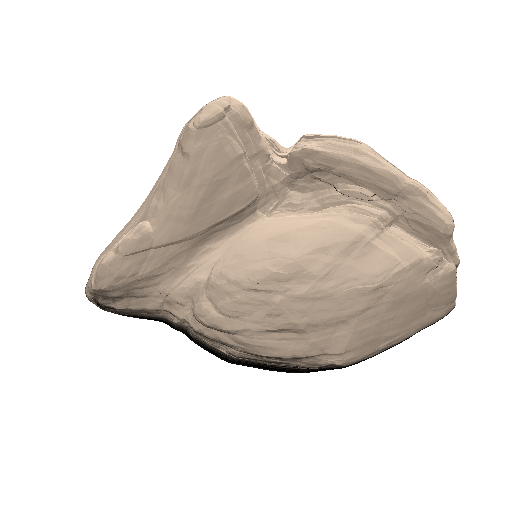} \\
    (a) Input & (b) with $\mathcal{L}_{\text{REG}}$ & (e) w/o $\mathcal{L}_{\text{REG}}$
    \end{tabular}
    \caption{\textbf{The effect of $\mathcal{L}_{\text{REG}}$.} Without the regularization term, our model fails to reconstruct 3D shapes from a sketch image.}
    \label{fig:reg}
\end{figure}

\paragraph{Regularization Loss.} We apply the same regularization loss as DualSDF~\cite{hao2020dualsdf}, \ie $\mathcal{L}_{\text{reg}} = \gamma \max(\|\*z\|^2_2, \beta)$, where two hyperparameters $\gamma$ and $\beta$ control the strength of the regularization. The regularization term $L_{\text{REG}}(\*z)$ effectively constrains the optimization of the latent code $\*z$ in the prior distribution of the pretrained MM-VADs.  Without such regularization, we find that the single-view 3D reconstruction fails in most cases. Fig.~\ref{fig:reg} provides one example. 

\paragraph{Multi-trial latent optimization for 3D reconstruction.} Similar to other generative models (\eg GANs), the latent optimization with the proposed MM-VADs is a highly non-convex problem and prone to local minimal. To relieve this issue, we conduct the latent optimization for multiple rounds with different initial latent codes. We use the latent codes with minimal reconstruction loss in multiple trials as the final results of the latent optimization. We find this simple strategy significantly stabilizes our model in the 3D reconstruction task. For example, the mean Chamfer distance decreases from 5.50 to 1.73 in the task of 3D reconstruction from single-view sketch on ShapeNet airplanes and from 9.10 to 4.70 on ShapeNet chairs. In 3D shape manipulation, the latent optimization starts from a known latent code corresponding to the target shape to be edited, and we only run the latent optimization once.

\section{Baselines}\label{sec:baselines}

Here we present more details about the baselines used in our experiments. 
\begin{itemize}
    \item\textbf{Encoder-Decoder Networks}~\cite{guillard2021sketch2mesh,remelli2020meshsdf}. This model is originally designed for predicting 3D shapes from sketches, followed by a shape refinement step based on differentiable rendering. We re-purpose this model to reconstruct 3D shapes from RGB images by simply modifying the input channels in the first convolutional layer. We use the official implementations with default hyperparameter settings~\footnote{\url{https://github.com/cvlab-epfl/MeshSDF}}. 
    \item \textbf{EditNeRF}~\cite{liu2021editing} edits a conditional radiance field representation of 3D scenes with sparse scribbles as input. The shape and color of 3D objects are edited by updating the neural network weights. We make qualitative comparisons with the EditNeRF using their pre-trained models\footnote{\url{https://github.com/stevliu/editnerf}}. Our model shares many similarities with EditNeRF (\eg network architecture, scribble-based interaction). However, the proposed model is significantly different from EditNeRF in terms of shape representation (SDFs~\cite{park2019deepsdf} vs NeRF~\cite{mildenhall2020nerf}), shape manipulation method (latent optimization vs network fine-tuning), and the way to bridge the 3D and 2D modalities (shared latent spaces vs differentiable rendering). Tab.~\ref{tab:editnerf} provides detailed comparisons between EditNeRF and our model. 
\end{itemize}

\begin{table}[ht!]
    \setlength{\tabcolsep}{3pt}
  \centering
    \caption{\textbf{Comparisons with EditNeRF}~\cite{liu2021editing}. $^\dag$ The shape reconstruction and manipulation can be combined and interleaved with the proposed model. This enables us to edit novel instances (Fig.~\ref{fig:real} in the main manuscript provides an example). $^\ddag$ The time cost of rendering a $256\times256$ image is included in the editing time}
   \begin{tabular}{r | cc}
    \toprule
     &  EditNeRF~\cite{liu2021editing} & Ours  \\
     \midrule
 Latent codes &  \multicolumn{2}{c}{Separate shape and color codes}  \\
 Network & \multicolumn{2}{c}{A common network shared by all training instances} \\ 
 Task & \multicolumn{2}{c}{Shape/color manipulation with sparse scribbles} \\ 
 \midrule 
  Instance-specific sub-networks &  \greencmark & \redxmark \\
  Generative model &  \redxmark &\greencmark \\
  3D recon. from sketch or RGB &  \redxmark &\greencmark \\
  Editing novel instances $^\dag$ &\redxmark &\greencmark \\
  Shape representation & NeRF~\cite{mildenhall2020nerf} & SDFs~\cite{park2019deepsdf} \\
  Bridge of 2D/3D modalities & Differentiable rendering & Shared latent spaces \\
  Editing method & Update network weights& Latent optimization \\
  Estimated editing time $^\ddag$ & ~60s &  ~7s \\
    \bottomrule
  \end{tabular}
  \label{tab:difference}
  \label{tab:editnerf}
    \vspace{-7pt}
\end{table}

\section{More Experimental details}\label{sec:more-details}

\subsection{Training and Testing Dataset}

We train the proposed multi-modal variational auto-decoders (MM-VADs) on the ShapeNet dataset~\cite{chang2015shapenet}. The training and testing split is the same as DeepSDF~\cite{park2019deepsdf} and DualSDF~\cite{hao2020dualsdf}. We use the same pre-trained MM-VADs throughout our experiments. For airplanes, there are 1780 shapes for training and 456 shapes for testing. For chairs, there are 3281 training shapes and 833 testing instances. For 3D shape manipulation, we present the results on known shapes (\ie shapes from training data), similar to EditNeRF~\cite{liu2021editing} and DualSDF~\cite{hao2020dualsdf}.  

\subsection{3D reconstruction from Sketch or RGB modalities.}

Table~\ref{tab:single-view} presents quantitative evaluations of the 3D reconstruction from sketch and RGB inputs under different occlusion ratios, corresponding to the curves in Fig.~\ref{fig:curves} in the main manuscript. We report results on both vertically and horizontally occluded inputs. Since 3D shapes and their 2D views are generally symmetric horizontally, the proposed model has almost no performance drop when masking out the right-half regions of the inputs. In comparison, the encoder-decoder networks~\cite{guillard2021sketch2mesh} that is trained on full-view inputs suffers from the input domain shift induced by the occlusion.

\begin{table}[ht!]
    \setlength{\tabcolsep}{9pt}
  \centering
    \caption{\label{tab:single-view}
    \textbf{Quantitative results of single-view reconstruction.} We report the average Chamfer Distance ($30,000$ points) multiplied by $10^3$ between the reconstructed 3D shapes and the groundtruth (\emph{lower} is \emph{better}). 
    The performance of the proposed model is slightly worse than the encoder-decoder networks~\cite{guillard2021sketch2mesh} trained on the full-view inputs. 
    However, MM-VADs perform more robustly to the input domain shift (\eg only partial view of input is available).
    The first column presents the occlusion rate in the input, where ``Full" means no occlusion in the input, ``1/2-horizontal" the left half of the input is visible, and ``3/4-vertical" the top 3/4 region of the object is available. 
    Superscripts in the last row denote the performance drop under the input domain shift (\emph{lower} is \emph{better}). 
    This table corresponds to Fig.~\ref{fig:curves} in the main text 
    }
  \begin{tabular}{c c rrrrrr}
    \toprule
  \multirow{2}{*}{View}    & \multirow{2}{*}{Model} & \multicolumn{2}{c}{Airplane} &  \multicolumn{2}{c}{Chair} & \multirow{2}{*}{Avg.} \\
  &  & Sketch & RGB & Sketch & RGB & \\
  \midrule
    \multirow{2}{*}{Full} &Enc-Dec  & \textbf{1.45} & \textbf{1.21} & \textbf{4.24} & \textbf{3.45}  & \textbf{2.59} \\
    &Ours & 1.73  & 1.40  & 5.96 & 4.70 & 3.44\\
      \midrule
     \multirow{2}{*}{1/2-horizontal} &Enc-Dec & 3.30  & 6.18 & 16.34 & 7.61 & 8.36\rlap{\red{+5.77}}\\
    &Ours & \textbf{1.79} & \textbf{1.38} & \textbf{6.07} & \textbf{5.00} & \textbf{3.56}\rlap{\red{+0.12}}\\
    \midrule
    \multirow{2}{*}{3/4-vertical} &Enc-Dec  & 2.33 & 1.94& 13.10 & 6.99 & 6.09\rlap{\red{+3.50}}\\
    &Ours & \textbf{2.07} & \textbf{1.55} & \textbf{6.91} & \textbf{5.64} & \textbf{4.04}\rlap{\red{+0.60}}\\
    \midrule
    \multirow{2}{*}{1/2-vertical} &Enc-Dec  & 3.97  & 3.56  & 24.31 & 10.13 & 10.49\rlap{\red{+7.90}}\\
    &Ours & \textbf{2.39} & \textbf{1.89} & \textbf{8.01} & \textbf{7.06} & \textbf{4.87}\rlap{\red{+1.43}}\\
    \midrule
    \multirow{2}{*}{1/4-vertical} &Enc-Dec  & 4.28 & 4.77 & 27.64 & 9.77 & 11.61\rlap{\red{+9.02}}\\
    &Ours & \textbf{3.32} & \textbf{2.63} & \textbf{8.27} & \textbf{8.19} & \textbf{5.60}\rlap{\red{+2.16}}\\
    \bottomrule
  \end{tabular}
\end{table}

\subsection{Few-shot 3D Generation}

\begin{figure}
    \centering
    \includegraphics[width=\linewidth]{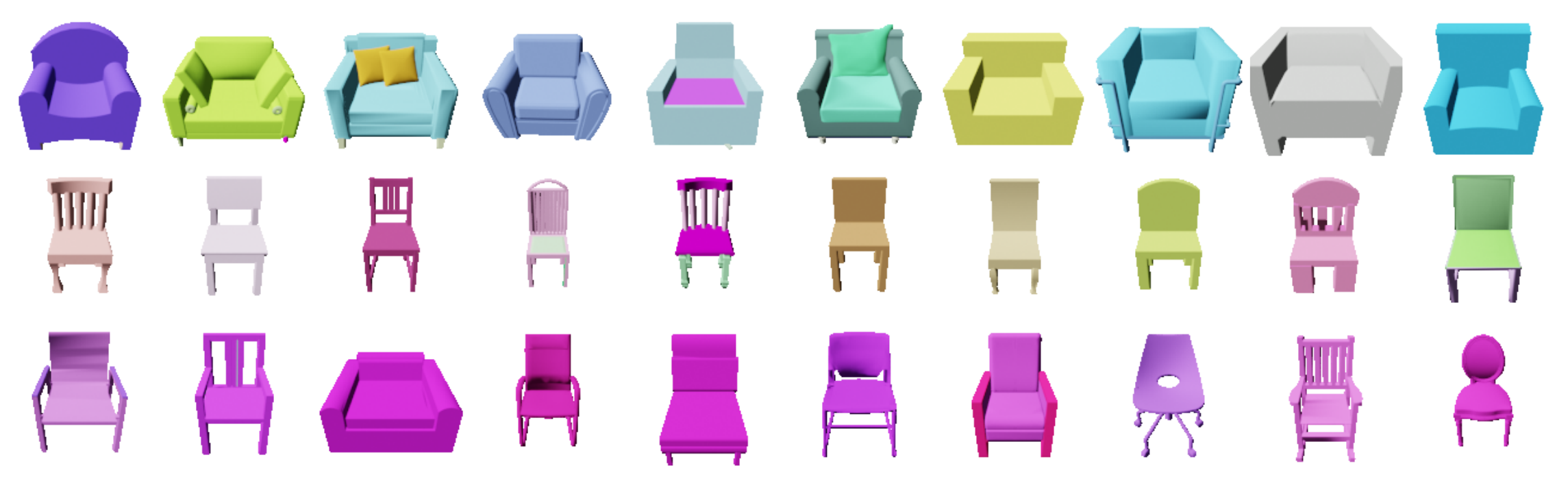}
    \caption{\textbf{2D examples for few-shot shape generation.} Each row presents the 10 2D examples used to adapt a pre-trained MM-VADs to generate armchairs, side chairs, and pink chairs respectively.}
    \label{fig:few-shot-examples}
\end{figure}

Fig.~\ref{fig:few-shot-examples} presents the 2D examples used in our few-shot shape generation experiments. For each category (\eg armchair), we randomly sample 10 images from our training data. We then adapt a pre-trained MM-VAD using these 2D examples based on the MineGAN framework~\cite{wang2020minegan}. We further collect 200 images per category from our training data for training binary classifiers and calculating FID scores. The classifiers are fine-tuned from a ResNet18~\cite{he2016deep} pre-trained on ImageNet. 

\section{Limitations}\label{sec:failure}

\begin{figure*}[ht!]
    \centering
    \begin{tabular}{c}
    \includegraphics[width=0.8\linewidth]{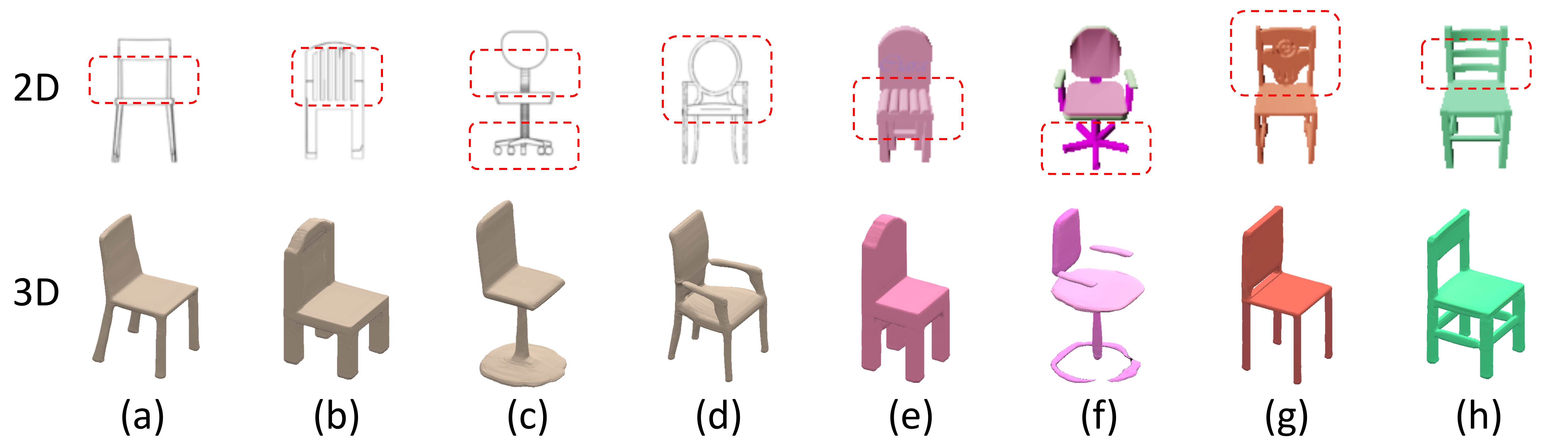} \\
    \end{tabular}
    \caption{\textbf{Limitations of 3D reconstruction from 2D modalities}. The proposed model fails to generate \emph{fine structures} of 3D shapes from sketches (a-d) or RGB renderings (e-h). The red bounding boxes highlight the object parts where our model fails to reconstruct the 3D structures.}
    \label{fig:failure-recon}
    \vspace{-7pt}
\end{figure*}

\paragraph{3D reconstruction from 2D modalities.} The proposed model fails to reconstruct \emph{fine structures} of 3D shapes from sketches or RGB views, for example, the holes on the back of chairs (Fig.~\ref{fig:failure-recon}a, b, g, h), fine textures on the seat of chairs (Fig.~\ref{fig:failure-recon}e), or the wheelbase of desk chairs (Fig.~\ref{fig:failure-recon}c, f). The capability of modeling fine structures is mainly determined by the 3D shape representation (\ie SDFs~\cite{park2019deepsdf}), training samples of SDFs, and the capacity of the proposed generative model. This issue can be potentially relieved by sampling more 3D training points surrounding the surface or increasing the capacity of the proposed model (\eg enlarging the dimension of the latent space, increasing the depth of 3D shape networks)

\begin{figure*}[ht!]
    \centering
    \begin{tabular}{c}
    \includegraphics[width=\linewidth]{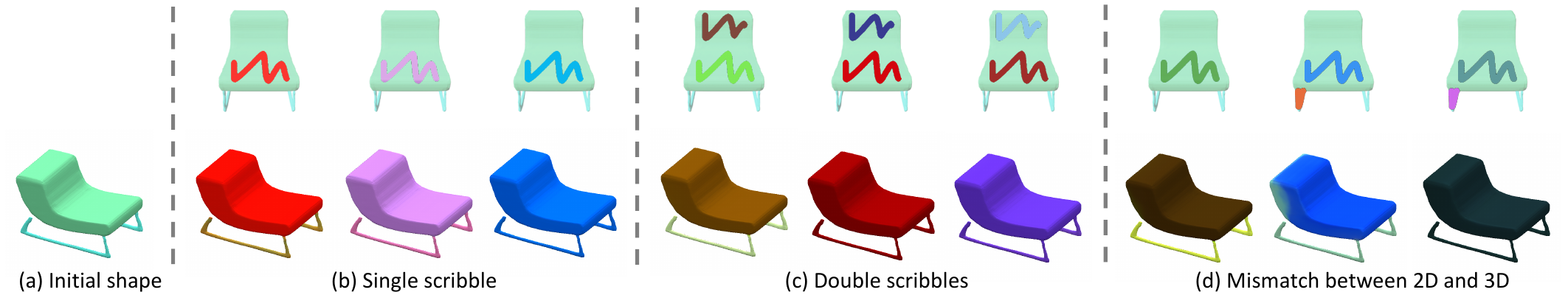} \\
    \end{tabular}
    \caption{\textbf{Limitations of editing shape via color scribble}. We are only able to provide color editing results in the prior distribution of the generative model. For example, if there are two scribbles of different color on the same part of a chair, our model either propagates one of the scribbles (\eg first two columns in (c)) or generates a surface color that are different from both scribbles (\eg last column in (c)). As a reference, we provide the editing results with single scribble in (b). Our model also produces 3D color editing results that do not match with the 2D input scribbles, as shown in (d).}
    \label{fig:failure-coloredit}
    \vspace{-7pt}
\end{figure*}

\paragraph{3D manipulation with 2D color scribble.} Similar to GAN-based image manipulation models~\cite{zhu2020domain,gu2020image,bau2020semantic,pan2020exploiting}, we are only able to provide editing results within the prior distribution of a pre-trained MM-VAD. For example, in the task of editing shape with color scribbles, if there are multiple scribbles of different colors on the same part of a shape (\eg the seat of a chair), our model either edits the shape based on one of the scribbles or generates a surface color that is completely different from all scribbles, as shown in Fig.~\ref{fig:failure-coloredit}. We notice that the editing results of EditNeRF~\cite{liu2021editing} are similar to ours based on their released demo\footnote{\url{https://github.com/stevliu/editnerf}}. 
Our model may produce unexpected color editing results, for example, the edited 3D surface color may not match the 2D color scribbles provided by the user (Fig.~\ref{fig:failure-coloredit}d), probably due to bad initialization of the latent code or suboptimal hyperparameter settings. The multi-trial latent optimization described in Sec.~\ref{sec:ablation} may relieve this issue.

\begin{figure*}[ht!]
    \centering
    \begin{tabular}{c}
    \includegraphics[width=\linewidth]{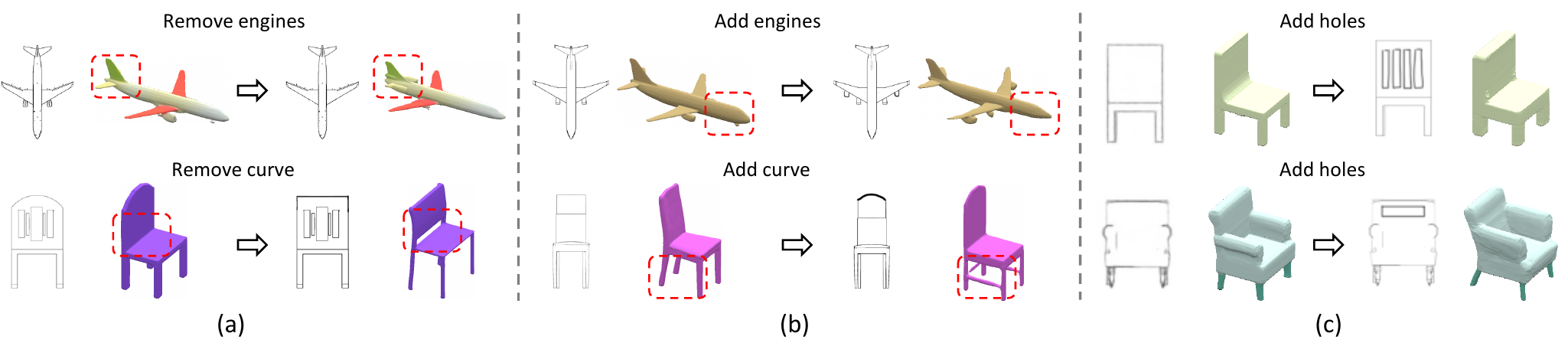} \\
    \end{tabular}
    \caption{\textbf{Limitations of editing shape via sketch}. \textbf{(a-b)} Editing one part via sketch leads to changes in other parts which are not edited in the sketch. The parts where our model fails to maintain are annotated in red bounding boxes. \textbf{(c)} The proposed method fails to add fine structures onto the shape via sketch (\eg adding holds onto the back of chairs).}
    \label{fig:failure-sketchedit}
\end{figure*}

\paragraph{3D manipulation via 2D sketch.} In this task, the major issue is that editing one part of a shape usually leads to changes in other parts. For example, removing the engines on the wing of airplanes results in new engines on the tail in many cases, as shown in Fig.~\ref{fig:sketch} in the main text. Fig.~\ref{fig:failure-sketchedit} in this section provides more examples. This is mainly because editing shapes via latent optimization can only produce new shapes in the prior distribution of the generative model. 
It is potentially useful to add more constraints upon the latent optimization, 
\eg enforcing the output of the 2D sketch generator to be as similar as possible to the original sketch. 
However, our preliminary experiments show that the latent optimization with such constraint typically under-fits the edited parts of the sketch and fails to achieve desired edits in 3D shape. 
In addition, the proposed model fails to add more complicated structures into the shape, for example, adding holes onto the back of chairs (Fig.~\ref{fig:failure-sketchedit}c). 
We will investigate these issues further in our future work.

\paragraph{Few-shot shape generation.} We are unable to adapt a pre-trained MM-VAD to generate shapes of \emph{fine-grained categories} (\eg single-engine airplanes) using a few 2D RGB images. We also fail to adapt a pre-trained MM-VAD using a few \emph{2D sketches}. We hypothesize that this is because the discriminator is trained from scratch and unable to learn discriminative representations among fine-grained categories or sparse inputs (\eg sketches) with limited 2D examples. 
These issues may be relieved by initializing the discriminator with a pre-trained classifier. 
We leave this in our future work.

\section{Diverse color scribbles.}\label{sec:scribble}
\begin{figure}[!h]
    \setlength{\tabcolsep}{1pt}
    \centering
    \vspace{-10pt}
    \begin{tabular}{c}
        \includegraphics[width=0.8\linewidth]{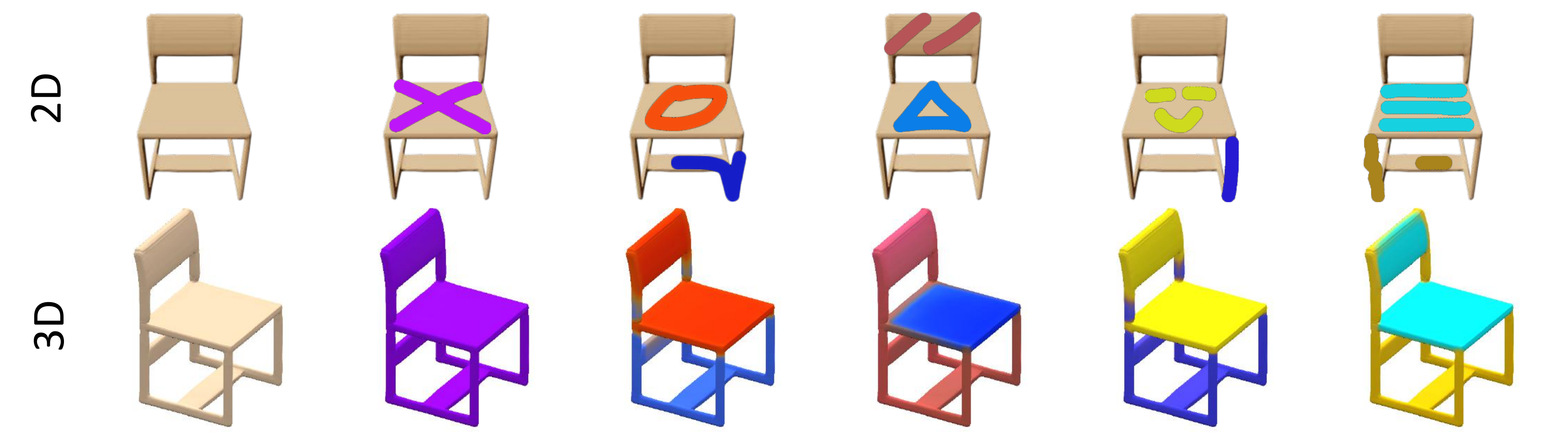}\\
       \end{tabular}
          \caption{\textbf{Diverse color scribbles}. The first column presents the initial 2D and 3D modalities. The following columns present the color editing results with diverse scribbles.}
    \label{fig:scribble-diverse}  
\end{figure}

 Fig.~\ref{fig:scribble-diverse} shows more 3D color editing results with diverse color scribbles. Our method is robust to color scribbles of different shapes/amounts/positions.  

%%%%%%%%% REFERENCES
% \clearpage
\bibliographystyle{splncs04}
\bibliography{reference}

\begin{thebibliography}{10}
\providecommand{\url}[1]{\texttt{#1}}
\providecommand{\urlprefix}{URL }
\providecommand{\doi}[1]{https://doi.org/#1}

\bibitem{abdal2019image2stylegan}
Abdal, R., Qin, Y., Wonka, P.: Image2stylegan: How to embed images into the
  stylegan latent space? In: Proceedings of the IEEE/CVF International
  Conference on Computer Vision. pp. 4432--4441 (2019)

\bibitem{an2011appwarp}
An, X., Tong, X., Denning, J.D., Pellacini, F.: Appwarp: Retargeting measured
  materials by appearance-space warping. In: Proceedings of the 2011 SIGGRAPH
  Asia Conference. pp. 1--10 (2011)

\bibitem{athar2018latent}
Athar, S., Burnaev, E., Lempitsky, V.: Latent convolutional models. In: ICLR
  (2018)

\bibitem{bau2020rewriting}
Bau, D., Liu, S., Wang, T., Zhu, J.Y., Torralba, A.: Rewriting a deep
  generative model. In: eccv. pp. 351--369. Springer (2020)

\bibitem{bau2020semantic}
Bau, D., Strobelt, H., Peebles, W., Zhou, B., Zhu, J.Y., Torralba, A., et~al.:
  Semantic photo manipulation with a generative image prior. arXiv preprint
  arXiv:2005.07727  (2020)

\bibitem{burt1987laplacian}
Burt, P.J., Adelson, E.H.: The laplacian pyramid as a compact image code. In:
  Readings in computer vision, pp. 671--679. Elsevier (1987)

\bibitem{chang2015shapenet}
Chang, A.X., Funkhouser, T., Guibas, L., Hanrahan, P., Huang, Q., Li, Z.,
  Savarese, S., Savva, M., Song, S., Su, H., et~al.: Shapenet: An
  information-rich 3d model repository. arXiv preprint arXiv:1512.03012  (2015)

\bibitem{chen2018text2shape}
Chen, K., Choy, C.B., Savva, M., Chang, A.X., Funkhouser, T., Savarese, S.:
  Text2shape: Generating shapes from natural language by learning joint
  embeddings. In: ACCV. pp. 100--116. Springer (2018)

\bibitem{chen2019learning}
Chen, Z., Zhang, H.: Learning implicit fields for generative shape modeling.
  In: Proceedings of the IEEE/CVF Conference on Computer Vision and Pattern
  Recognition. pp. 5939--5948 (2019)

\bibitem{choi2018stargan}
Choi, Y., Choi, M., Kim, M., Ha, J.W., Kim, S., Choo, J.: Stargan: Unified
  generative adversarial networks for multi-domain image-to-image translation.
  In: Proceedings of the IEEE conference on computer vision and pattern
  recognition. pp. 8789--8797 (2018)

\bibitem{choy20163d}
Choy, C.B., Xu, D., Gwak, J., Chen, K., Savarese, S.: 3d-r2n2: A unified
  approach for single and multi-view 3d object reconstruction. In: European
  conference on computer vision. pp. 628--644. Springer (2016)

\bibitem{decarlo2003suggestive}
DeCarlo, D., Finkelstein, A., Rusinkiewicz, S., Santella, A.: Suggestive
  contours for conveying shape. In: ACM SIGGRAPH 2003 Papers, pp. 848--855
  (2003)

\bibitem{delanoy20183d}
Delanoy, J., Aubry, M., Isola, P., Efros, A.A., Bousseau, A.: 3d sketching
  using multi-view deep volumetric prediction. Proceedings of the ACM on
  Computer Graphics and Interactive Techniques  \textbf{1}(1),  1--22 (2018)

\bibitem{fan2017point}
Fan, H., Su, H., Guibas, L.J.: A point set generation network for 3d object
  reconstruction from a single image. In: Proceedings of the IEEE conference on
  computer vision and pattern recognition. pp. 605--613 (2017)

\bibitem{gkioxari2019mesh}
Gkioxari, G., Malik, J., Johnson, J.: Mesh r-cnn. In: Proceedings of the
  IEEE/CVF International Conference on Computer Vision. pp. 9785--9795 (2019)

\bibitem{goel2020shape}
Goel, S., Kanazawa, A., Malik, J.: Shape and viewpoint without keypoints. In:
  European Conference on Computer Vision. pp. 88--104. Springer (2020)

\bibitem{goodfellow2014generative}
Goodfellow, I., Pouget-Abadie, J., Mirza, M., Xu, B., Warde-Farley, D., Ozair,
  S., Courville, A., Bengio, Y.: Generative adversarial nets. Advances in
  neural information processing systems  \textbf{27} (2014)

\bibitem{grady2006random}
Grady, L.: Random walks for image segmentation. IEEE transactions on pattern
  analysis and machine intelligence  \textbf{28}(11),  1768--1783 (2006)

\bibitem{gu2020image}
Gu, J., Shen, Y., Zhou, B.: Image processing using multi-code gan prior. In:
  Proceedings of the IEEE/CVF Conference on Computer Vision and Pattern
  Recognition. pp. 3012--3021 (2020)

\bibitem{guillard2021sketch2mesh}
Guillard, B., Remelli, E., Yvernay, P., Fua, P.: Sketch2mesh: Reconstructing
  and editing 3d shapes from sketches. In: ICCV (2021)

\bibitem{gulrajani2017improved}
Gulrajani, I., Ahmed, F., Arjovsky, M., Dumoulin, V., Courville, A.: Improved
  training of wasserstein gans. In: NeurIPS (2017)

\bibitem{hao2020dualsdf}
Hao, Z., Averbuch-Elor, H., Snavely, N., Belongie, S.: Dualsdf: Semantic shape
  manipulation using a two-level representation. In: CVPR. pp. 7631--7641
  (2020)

\bibitem{he2016deep}
He, K., Zhang, X., Ren, S., Sun, J.: Deep residual learning for image
  recognition. In: Proceedings of the IEEE conference on computer vision and
  pattern recognition. pp. 770--778 (2016)

\bibitem{heusel2017gans}
Heusel, M., Ramsauer, H., Unterthiner, T., Nessler, B., Hochreiter, S.: Gans
  trained by a two time-scale update rule converge to a local nash equilibrium.
  Advances in neural information processing systems  \textbf{30} (2017)

\bibitem{ioffe2015batch}
Ioffe, S., Szegedy, C.: Batch normalization: Accelerating deep network training
  by reducing internal covariate shift. In: International conference on machine
  learning. pp. 448--456. PMLR (2015)

\bibitem{jin2020contour}
Jin, A., Fu, Q., Deng, Z.: Contour-based 3d modeling through joint embedding of
  shapes and contours. In: Symposium on Interactive 3D Graphics and Games. pp.
  1--10 (2020)

\bibitem{kanazawa2018learning}
Kanazawa, A., Tulsiani, S., Efros, A.A., Malik, J.: Learning category-specific
  mesh reconstruction from image collections. In: Proceedings of the European
  Conference on Computer Vision (ECCV). pp. 371--386 (2018)

\bibitem{kingma2014adam}
Kingma, D.P., Ba, J.: Adam: A method for stochastic optimization. arXiv
  preprint arXiv:1412.6980  (2014)

\bibitem{kingma2013auto}
Kingma, D.P., Welling, M.: Auto-encoding variational bayes. arXiv preprint
  arXiv:1312.6114  (2013)

\bibitem{lempitsky2009image}
Lempitsky, V., Kohli, P., Rother, C., Sharp, T.: Image segmentation with a
  bounding box prior. In: 2009 IEEE 12th international conference on computer
  vision. pp. 277--284. IEEE (2009)

\bibitem{levin2004colorization}
Levin, A., Lischinski, D., Weiss, Y.: Colorization using optimization. In: ACM
  SIGGRAPH 2004 Papers, pp. 689--694 (2004)

\bibitem{li2004lazy}
Li, Y., Sun, J., Tang, C.K., Shum, H.Y.: Lazy snapping. ACM Transactions on
  Graphics (ToG)  \textbf{23}(3),  303--308 (2004)

\bibitem{liu2016coupled}
Liu, M.Y., Tuzel, O.: Coupled generative adversarial networks. Advances in
  neural information processing systems  \textbf{29},  469--477 (2016)

\bibitem{liu2020dist}
Liu, S., Zhang, Y., Peng, S., Shi, B., Pollefeys, M., Cui, Z.: Dist: Rendering
  deep implicit signed distance function with differentiable sphere tracing.
  In: CVPR. pp. 2019--2028 (2020)

\bibitem{liu2019soft}
Liu, S., Li, T., Chen, W., Li, H.: Soft rasterizer: A differentiable renderer
  for image-based 3d reasoning. In: Proceedings of the IEEE/CVF International
  Conference on Computer Vision. pp. 7708--7717 (2019)

\bibitem{liu2021editing}
Liu, S., Zhang, X., Zhang, Z., Zhang, R., Zhu, J.Y., Russell, B.: Editing
  conditional radiance fields. In: ICCV (2021)

\bibitem{mescheder2019occupancy}
Mescheder, L., Oechsle, M., Niemeyer, M., Nowozin, S., Geiger, A.: Occupancy
  networks: Learning 3d reconstruction in function space. In: Proceedings of
  the IEEE/CVF Conference on Computer Vision and Pattern Recognition. pp.
  4460--4470 (2019)

\bibitem{mildenhall2020nerf}
Mildenhall, B., Srinivasan, P.P., Tancik, M., Barron, J.T., Ramamoorthi, R.,
  Ng, R.: Nerf: Representing scenes as neural radiance fields for view
  synthesis. In: European conference on computer vision. pp. 405--421. Springer
  (2020)

\bibitem{pan2020exploiting}
Pan, X., Zhan, X., Dai, B., Lin, D., Loy, C.C., Luo, P.: Exploiting deep
  generative prior for versatile image restoration and manipulation. In:
  European Conference on Computer Vision. pp. 262--277. Springer (2020)

\bibitem{park2019deepsdf}
Park, J.J., Florence, P., Straub, J., Newcombe, R., Lovegrove, S.: Deepsdf:
  Learning continuous signed distance functions for shape representation. In:
  CVPR. pp. 165--174 (2019)

\bibitem{pellacini2007lighting}
Pellacini, F., Battaglia, F., Morley, R.K., Finkelstein, A.: Lighting with
  paint. ACM Transactions on Graphics (TOG)  \textbf{26}(2),  9--es (2007)

\bibitem{radford2015unsupervised}
Radford, A., Metz, L., Chintala, S.: Unsupervised representation learning with
  deep convolutional generative adversarial networks. arXiv preprint
  arXiv:1511.06434  (2015)

\bibitem{remelli2020meshsdf}
Remelli, E., Lukoianov, A., Richter, S.R., Guillard, B., Bagautdinov, T.,
  Baque, P., Fua, P.: Meshsdf: Differentiable iso-surface extraction. arXiv
  preprint arXiv:2006.03997  (2020)

\bibitem{rother2004grabcut}
Rother, C., Kolmogorov, V., Blake, A.: " grabcut" interactive foreground
  extraction using iterated graph cuts. ACM transactions on graphics (TOG)
  \textbf{23}(3),  309--314 (2004)

\bibitem{saharia2021palette}
Saharia, C., Chan, W., Chang, H., Lee, C.A., Ho, J., Salimans, T., Fleet, D.J.,
  Norouzi, M.: Palette: Image-to-image diffusion models. arXiv preprint
  arXiv:2111.05826  (2021)

\bibitem{schmidt2016state}
Schmidt, T.W., Pellacini, F., Nowrouzezahrai, D., Jarosz, W., Dachsbacher, C.:
  State of the art in artistic editing of appearance, lighting and material.
  In: Computer Graphics Forum. vol.~35, pp. 216--233. Wiley Online Library
  (2016)

\bibitem{shen2020interpreting}
Shen, Y., Gu, J., Tang, X., Zhou, B.: Interpreting the latent space of gans for
  semantic face editing. In: Proceedings of the IEEE/CVF Conference on Computer
  Vision and Pattern Recognition. pp. 9243--9252 (2020)

\bibitem{shen2020interfacegan}
Shen, Y., Yang, C., Tang, X., Zhou, B.: Interfacegan: Interpreting the
  disentangled face representation learned by gans. IEEE Transactions on
  Pattern Analysis and Machine Intelligence  (2020)

\bibitem{shi2019variational}
Shi, Y., Siddharth, N., Paige, B., Torr, P.H.: Variational mixture-of-experts
  autoencoders for multi-modal deep generative models. arXiv preprint
  arXiv:1911.03393  (2019)

\bibitem{sitzmann2019scene}
Sitzmann, V., Zollh{\"o}fer, M., Wetzstein, G.: Scene representation networks:
  Continuous 3d-structure-aware neural scene representations. arXiv preprint
  arXiv:1906.01618  (2019)

\bibitem{sohl2015deep}
Sohl-Dickstein, J., Weiss, E., Maheswaranathan, N., Ganguli, S.: Deep
  unsupervised learning using nonequilibrium thermodynamics. In: ICML. pp.
  2256--2265. PMLR (2015)

\bibitem{suzuki2016joint}
Suzuki, M., Nakayama, K., Matsuo, Y.: Joint multimodal learning with deep
  generative models. arXiv preprint arXiv:1611.01891  (2016)

\bibitem{tatarchenko2019single}
Tatarchenko, M., Richter, S.R., Ranftl, R., Li, Z., Koltun, V., Brox, T.: What
  do single-view 3d reconstruction networks learn? In: CVPR. pp. 3405--3414
  (2019)

\bibitem{wang2018pixel2mesh}
Wang, N., Zhang, Y., Li, Z., Fu, Y., Liu, W., Jiang, Y.G.: Pixel2mesh:
  Generating 3d mesh models from single rgb images. In: ECCV. pp. 52--67 (2018)

\bibitem{wang2020minegan}
Wang, Y., Gonzalez-Garcia, A., Berga, D., Herranz, L., Khan, F.S., Weijer,
  J.v.d.: Minegan: effective knowledge transfer from gans to target domains
  with few images. In: CVPR. pp. 9332--9341 (2020)

\bibitem{wu2018multimodal}
Wu, M., Goodman, N.: Multimodal generative models for scalable
  weakly-supervised learning. Advances in Neural Information Processing Systems
   \textbf{31} (2018)

\bibitem{wu2019multimodal}
Wu, M., Goodman, N.: Multimodal generative models for compositional
  representation learning. arXiv preprint arXiv:1912.05075  (2019)

\bibitem{xu2019disn}
Xu, Q., Wang, W., Ceylan, D., Mech, R., Neumann, U.: Disn: Deep implicit
  surface network for high-quality single-view 3d reconstruction. arXiv
  preprint arXiv:1905.10711  (2019)

\bibitem{yang2021lasr}
Yang, G., Sun, D., Jampani, V., Vlasic, D., Cole, F., Chang, H., Ramanan, D.,
  Freeman, W.T., Liu, C.: Lasr: Learning articulated shape reconstruction from
  a monocular video. In: Proceedings of the IEEE/CVF Conference on Computer
  Vision and Pattern Recognition. pp. 15980--15989 (2021)

\bibitem{zadeh2019variational}
Zadeh, A., Lim, Y.C., Liang, P.P., Morency, L.P.: Variational auto-decoder: A
  method for neural generative modeling from incomplete data. arXiv preprint
  arXiv:1903.00840  (2019)

\bibitem{zhang2018perceptual}
Zhang, R., Isola, P., Efros, A.A., Shechtman, E., Wang, O.: The unreasonable
  effectiveness of deep features as a perceptual metric. In: CVPR (2018)

\bibitem{zhang2017real}
Zhang, R., Zhu, J.Y., Isola, P., Geng, X., Lin, A.S., Yu, T., Efros, A.A.:
  Real-time user-guided image colorization with learned deep priors. ACM
  Transactions on Graphics (TOG)  \textbf{9}(4) (2017)

\bibitem{zhang2021sketch2model}
Zhang, S.H., Guo, Y.C., Gu, Q.W.: Sketch2model: View-aware 3d modeling from
  single free-hand sketches. In: Proceedings of the IEEE/CVF Conference on
  Computer Vision and Pattern Recognition. pp. 6012--6021 (2021)

\bibitem{zhong2020deep}
Zhong, Y., Gryaditskaya, Y., Zhang, H., Song, Y.Z.: Deep sketch-based modeling:
  Tips and tricks. In: 2020 International Conference on 3D Vision (3DV). pp.
  543--552. IEEE (2020)

\bibitem{zhu2020domain}
Zhu, J., Shen, Y., Zhao, D., Zhou, B.: In-domain gan inversion for real image
  editing. In: ECCV. pp. 592--608. Springer (2020)

\end{thebibliography}
\end{document}